\documentclass[pdflatex,sn-mathphys-num]{sn-jnl}

\usepackage{graphicx}%
\usepackage{multirow}%
\usepackage{amsmath,amssymb,amsfonts}%
\usepackage{amsthm}%
\usepackage{mathrsfs}%
\usepackage[title]{appendix}%
\usepackage{xcolor}%
\usepackage{textcomp}%
\usepackage{manyfoot}%
\usepackage{booktabs}%
\usepackage{algorithm}%
\usepackage{algorithmicx}%
\usepackage{algpseudocode}%
\usepackage{listings}%

\usepackage{morefloats}

\usepackage{wrapfig}   

\usepackage{lineno}

\theoremstyle{thmstyleone}%
\theoremstyle{thmstyletwo}%

\theoremstyle{thmstylethree}%

\begin{document}

\title[Article Title]{How Deep is Love in LLMs' Hearts? Exploring Semantic Size in Human-like Cognition}


\author[1,2,3]{\fnm{Yao} \sur{Yao}}\email{yaoyao27@sjtu.edu.cn}

\author[1,2,3]{\fnm{Yifei} \sur{Yang}}\email{yifeiyang@sjtu.edu.cn}

\author[1,2,3]{\fnm{Xinbei} \sur{Ma}}\email{sjtumaxb@sjtu.edu.cn}

\author[1,2,3]{\fnm{Dongjie} \sur{Yang}}\email{djyang.tony@sjtu.edu.cn}

\author[4]{\fnm{Zhuosheng} \sur{Zhang}}\email{zhangzs@sjtu.edu.cn}

\author*[5]{\fnm{Zuchao} \sur{Li}}\email{zcli-charlie@whu.edu.cn}

\author*[1,2,3]{\fnm{Hai} \sur{Zhao}}\email{zhaohai@cs.sjtu.edu.cn}

\affil[1]{\orgdiv{Department of Computer Science and Engineering}, \orgname{Shanghai Jiao Tong University}, \orgaddress{\city{Shanghai}, \country{P. R. China}}}

\affil[2]{\orgdiv{Shanghai Key Laboratory of Trusted Data Circulation and Governance in Web3}, \orgaddress{\city{Shanghai}, \country{P. R. China}}}

\affil[3]{\orgdiv{Key Laboratory of Shanghai Education Commission for Intelligent Interaction and Cognitive Engineering}, \orgname{Shanghai Jiao Tong University}, \orgaddress{\city{Shanghai}, \country{P. R. China}}}

\affil[4]{\orgdiv{School of Electronic Information and Electrical Engineering}, \orgname{Shanghai Jiao Tong University}, \orgaddress{\city{Shanghai}, \country{P. R. China}}}

\affil[5]{\orgdiv{National Engineering Research Center for Multimedia Software}, \orgname{School of Computer Science, Wuhan University, Wuhan University}, \orgaddress{\postcode{430072},\city{Wuhan}, \country{P. R. China}}}

\abstract{How human cognitive abilities are formed has long captivated researchers. However, a significant challenge lies in developing meaningful methods to measure these complex processes. With the advent of large language models (LLMs), which now rival human capabilities in various domains, we are presented with a unique testbed to investigate human cognition through a new lens. Among the many facets of cognition, one particularly crucial aspect is the concept of semantic size—the perceived magnitude of both abstract and concrete words or concepts. 
This study seeks to investigate whether LLMs exhibit similar tendencies in understanding semantic size, thereby providing insights into the underlying mechanisms of human cognition. We begin by exploring metaphorical reasoning, comparing how LLMs and humans associate abstract words with concrete objects of varying sizes. Next, we examine LLMs' internal representations to evaluate their alignment with human cognitive processes.
Our findings reveal that multi-modal training is crucial for LLMs to achieve more human-like understanding, suggesting that real-world, multi-modal experiences are similarly vital for human cognitive development. 
Lastly, we examine whether LLMs are influenced by attention-grabbing headlines with larger semantic sizes in a real-world web shopping scenario. The results show that multi-modal LLMs are more emotionally engaged in decision-making, but this also introduces potential biases, such as the risk of manipulation through clickbait headlines. Ultimately, this study offers a novel perspective on how LLMs interpret and internalize language, from the smallest concrete objects to the most profound abstract concepts like `love'. The insights gained not only improve our understanding of LLMs but also provide new avenues for exploring the cognitive abilities that define human intelligence. Our code and dataset is publicly available at \href{https://github.com/Zoeyyao27/LLM-Semantic-Size-Understanding}{https://github.com/Zoeyyao27/LLM-Semantic-Size-Understanding}. 
}




\maketitle

\textit{Love is like a galaxy, vast and infinite, filled with countless stars of joy, challenges, and mysteries, each one contributing to the beauty and wonder of the whole, yet bound together by unseen forces. --- ChatGPT 4o}


\section{Introduction}\label{sec1}

Understanding human cognition, especially how it develops and operates, has long captivated researchers. One of the greatest challenges in this field lies in finding suitable methods to measure the intricacies of cognitive processes. With the rise of large language models (LLMs), we are presented with a unique opportunity to study cognition in ways that were previously unattainable. These models, which now rival human performance in domains such as reading comprehension, competition-level mathematics, and scientific question answering \cite{luo2024large, zhangmultimodal, yax2024studying, perrault2024artificial,yao2023beyond}, offer a valuable testbed for exploring the complexities of human cognitive abilities.

The impressive capabilities of LLMs stem from their immense scale and drawn widespread attention from researchers, igniting rapid advancements in the Artificial General Intelligence (AGI). Biological research indicates that the human brain contains approximately 86 billion neurons \cite{azevedo2009equal}, while current LLMs, such as GPT-3, have reached a similar scale, with over 100 billion parameters \cite{brown2020language}. This suggests that LLMs are approaching, and in some ways, surpassing the scale of the human brain's neural network \cite{zh_nlp}. Furthermore, similarities have shown between human language learning and LLM training. Studies \cite{kuhl2004early, brown1973first} propose that language acquisition is not a linear process but occurs in stages. Early language learning (e.g., infants’ perception of phonemes and basic mastery of their native language) can be viewed as a form of `pre-training', which lays the foundation for later learning of grammar and complex structures. 
For instance, children master the basic framework of their native language around age 4, refining these abilities through further input, much like the pre-training and fine-tuning phases of LLMs. 
Furthermore, LLMs have evolved from their original single-modality form (text only) to Multi-modal Large Language Models (MLLMs) \cite{wu2023next, achiam2023gpt, bai2023qwen_vl,young2024yi,yao2024minicpm} capable of processing modalities including text, audio, video, and images, adding `senses' to a language-based `brain' enables multi-modal agents to act in a human-like manner.
These parallels between LLMs and the human brain have begun to attract researchers' attention, prompting exploration into the similarities and differences between the two \cite{yax2024studying, moro2023large, kumar2024shared}.

One particularly crucial aspect of cognition, which offers a window into broader cognitive mechanisms, is the concept of semantic size \cite{majid2004can, barsalou1999perceptual}. 
Semantic size measures the magnitude of a concept or object, categorized as big or small, in both concrete and abstract terms \cite{scott2019glasgow}. For instance, words such as `spaceship' or `love' convey larger semantic sizes, while terms like `dust' or `hush' suggest smaller sizes. A more detailed example of different words and their semantic sizes can be seen in Figure 1, which illustrates the semantic size of various words based on the Glasgow Norms dataset \cite{scott2019glasgow} \footnote{The numbers beneath each word represent the word's score on the SIZE dimension from the Glasgow Norms. More details will be discussed in section \ref{sec:glasgow}.}. The understanding of semantic size is essential in structuring how humans process and categorize information, and examining how LLMs handle this concept may provide new insights into both artificial and human cognition. 

Research has shown that semantic size significantly influences conceptual processing, particularly in concrete concepts where physical size can be measured, underscoring the pivotal role that semantic size plays in shaping human cognition \cite{rubinsten2002ant}. 
\begin{wrapfigure}{r}{0.5\textwidth}
    \centering
    \includegraphics[width=0.4\textwidth]{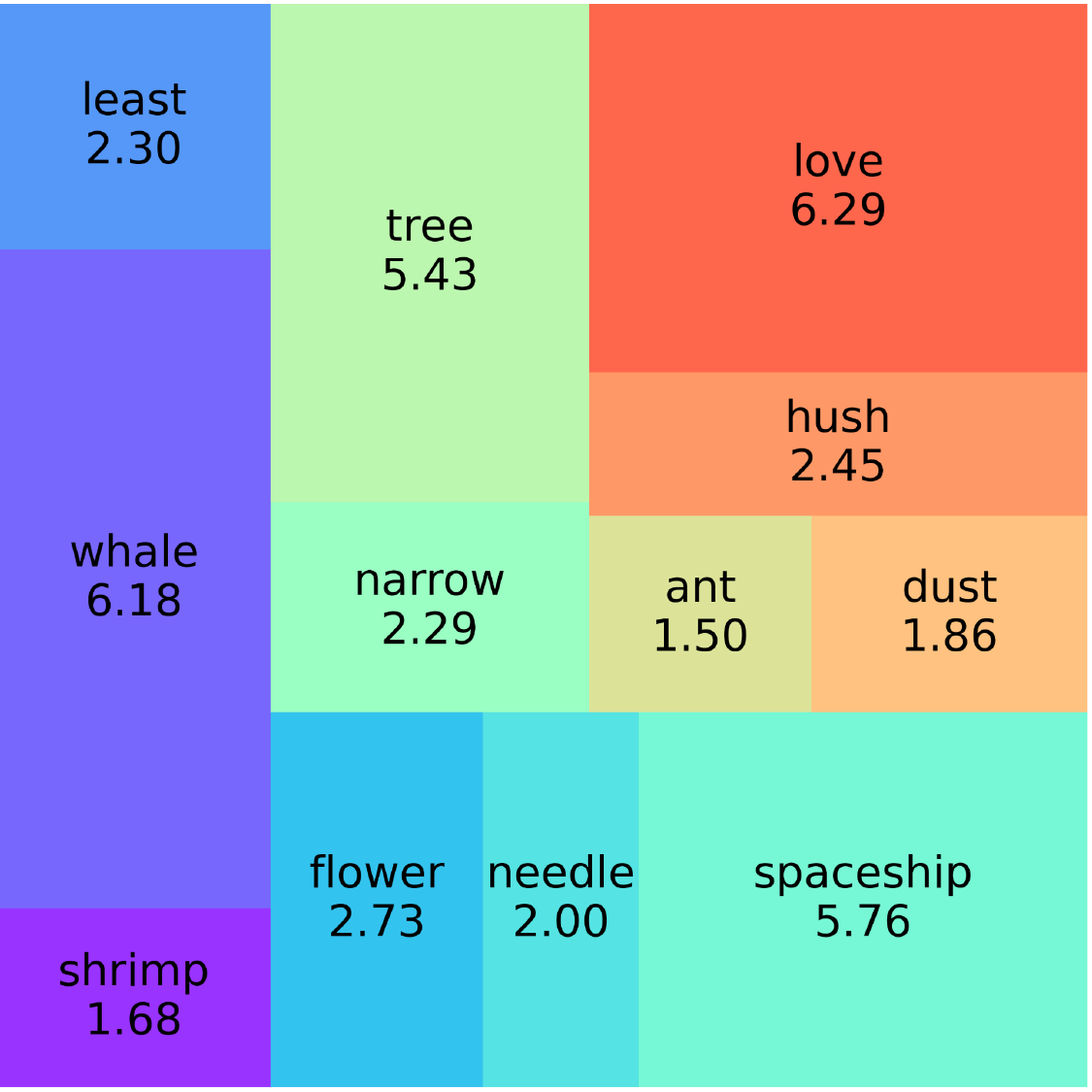}
    \caption{The image displays the semantic size of various words, derived from the Glasgow Norms dataset \cite{scott2019glasgow} . The area of each block represents the relative semantic size of the corresponding word, with larger blocks indicating greater semantic size.}
    \label{fig:treemap}
\end{wrapfigure} 
Consequently, numerous studies have examined how the human brain perceives the semantic size of words \cite{yao2022can, sereno2009short, yao2013semantic, pinel2004distributed}. 
However, the question of how LLMs understand semantic size remains under-explored. Research has shown that people's understanding of the semantic size of both abstract and concrete concepts is rooted in their multi-modal experiences \cite{Barsalou_1999, Barsalou_2008}. For humans, embodied experiences significantly contribute to the semantic size of words, with the understanding of abstract words proving more challenging than that of concrete words \cite{yao2022can, schwanenflugel2013abstract}. This raises the question: can LLMs, lacking embodied experiences, comprehend the semantic size of words in the same way humans do? Furthermore, do multi-modal trained MLLMs, equipped with visual grounding, enhance their ability to understand semantic size?\footnote{In this study, we primarily focus on MLLMs trained with visual modality.} In addition, based on the similarities between LLMs and the human brain, the conclusions drawn from LLM studies may provide indirect insights into how human cognitive abilities develop.

\begin{figure}
    \centering
    \includegraphics[width=1\linewidth]{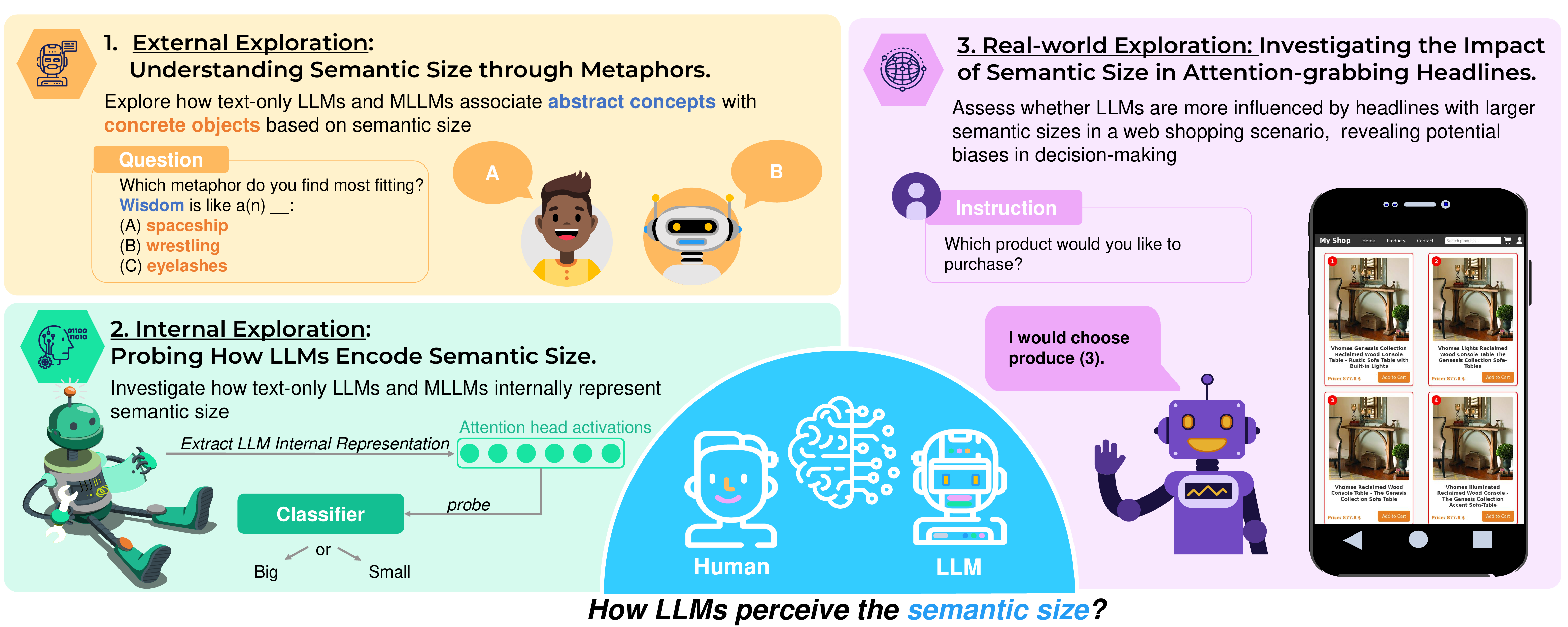}
    \caption{ Overview of how LLMs perceive semantic size from three perspectives. External exploration examines metaphorical associations between abstract concepts and concrete objects, comparing LLMs and human cognition. Internal exploration probes how LLMs encode and represent semantic size. Real-world exploration assesses how semantic size influences LLMs’ responses to attention-grabbing headlines in a web shopping scenario, revealing potential biases in decision-making.}
    \label{fig:overview}
\end{figure}

Therefore, we aim to venture into this uncharted territory, seeking to understand how LLMs interpret the semantic size of words in comparison to human cognition. In this paper, we progressively explore how LLMs perceive the semantic size of language from three perspectives: (1) External Exploration: Understanding Semantic Size through Metaphors, (2) Internal Exploration: Probing How LLMs Encode Semantic Size, and (3) Real-world Exploration: Investigating the Impact of Semantic Size in Attention-grabbing Headlines. An overview of the three computational studies is presented in Figure \ref{fig:overview}. We will now briefly walk through each of these three computational studies.

\textbf{1) External Exploration: Understanding Semantic Size through Metaphors.} We begin by using a metaphor test to explore LLMs' understanding of semantic size. Metaphors serve as powerful tools in semantics \cite{Cormac1990-CORACT-4}. The human ability to create metaphors stems from a deep understanding of the semantics of the words being compared. Some studies have shown that for humans, the semantic size of abstract words is metaphorically associated with the physical size of concrete objects \cite{yao2022can}. Building on this, we categorize words into abstract and concrete groups, asking LLMs to associate abstract words with concrete words according to their semantic size. This allows us to explore whether LLMs exhibit the same metaphorical tendencies as humans, providing insight into whether LLMs can understand not only concrete words like `spaceship' but also more complex abstract concepts like `love' from the novel perspective of semantic size. 

Our results indicate that some LLMs struggle to generate effective metaphors from a semantic-size perspective, while \textbf{MLLMs tend to align more closely with human metaphorical tendencies compared to their backbone text-only LLMs counterparts}.

\textbf{2) Internal Exploration: Probing How LLMs Encode Semantic Size.} Following our metaphor-based investigation, we turn to a more technical approach, examining the internal representations of LLMs to evaluate whether they encode semantic size effectively. We begin by training linear classifier probes on the latent representations within the language model to predict the semantic size of a given word. This probing technique allows us to assess how well LLMs capture semantic size in their internal structure.


Our findings suggest that MLLMs outperform single-modality LLMs in semantic size classification accuracy, indicating that \textbf{multi-modality plays a critical role in equipping LLMs with a more comprehensive cognitive capacity.} Furthermore, through these first two computational studies of LLMs,\textbf{ we can indirectly demonstrate that human cognition cannot be fully acquired through the language modality alone. In other words, both LLMs and the human brain require real-world engagement and multi-modal experiences as essential pathways to develop comprehensive cognitive abilities.}

\textbf{3) Real-world Exploration: Investigating the Impact of Semantic Size in Attention-grabbing Headlines.}
Previous research has demonstrated that semantic size is closely linked to emotional arousal \cite{yao2013semantic}, and attention-grabbing headlines are known to generate higher levels of emotional arousal compared to neutral headlines \cite{chen2015misleading, pengnate2019shocking}. Building on this, we investigate whether LLMs exhibit a bias toward attention-grabbing headlines with larger semantic size. Specifically, we ask the question: `Are LLMs more likely to be influenced by attention-grabbing headlines with a larger semantic size?'

To address this, we constructed a multi-modal dataset simulating a shopping website. We refined product titles with three different prompts, each designed to increase the semantic size of the original titles to varying degrees. We calculated the average semantic size of the resulting product titles and found that, when all other factors were controlled, \textbf{LLMs, especially MLLMs, showed a clear preference for selecting product titles with larger semantic sizes. This suggests that LLMs, much like humans, are more attracted to attention-grabbing headlines when its semantic size is larger, revealing a tendency toward content that is perceived as more emotionally engaging.}

Through these three explorations, we aim to uncover how different LLMs comprehend semantic size and whether they exhibit human-like tendencies in understanding semantic size, offering insights into the underlying mechanisms of human cognition. By focusing on semantic size, this progression from abstract associations to practical applications allows us to evaluate not only how effectively LLMs replicate human cognitive processes, but also to glimpse the broader cognitive structures that shape both artificial and human intelligence.

\section{External Exploration: Understanding Semantic Size through Metaphors.}
\label{sec:ex}
With the continuous advancements in language and cognitive science, metaphors have become central to human abstract conceptualization and reasoning. This suggests that metaphors are not merely rhetorical tools but are deeply embedded in human cognition. They enable us to understand and navigate complex, abstract concepts by drawing on more tangible, concrete references \cite{Johnson2010}.

Recent studies have revealed that humans metaphorically associate the semantic size of abstract concepts with the physical size of concrete objects. For example, larger abstract concepts (e.g., `love') are often linked to larger physical objects (e.g., `spaceship'), while smaller abstract concepts (e.g., `hush') are associated with smaller objects (e.g., `dust') \cite{yao2022can}.  This phenomenon is thought to arise because abstract concepts, unlike concrete ones, lack direct physical referents and to help mentally represent and understand these abstract ideas, humans often rely on metaphors drawn from their embodied experiences with the physical world. This process of grounding abstract concepts in concrete experiences helps make these intangible ideas more relatable and comprehensible. For instance, concepts like `love' are often associated with physical objects that are large because love is metaphorically perceived as vast and powerful, while concepts like `hush' are linked to smaller objects as they are perceived as minor or delicate. These metaphorical associations occur because our cognitive processing system tends to utilize size as a way of representing not only physical magnitude but also emotional or conceptual importance. The study supports this with evidence showing that abstract size can influence conceptual processing in similar ways to physical size \cite{yao2013semantic,yao2022can}. This connection between semantic size and metaphor offers a unique window into how humans conceptualize abstract ideas.

Given that single-modality LLMs are trained exclusively on text and lack embodied experiences with the physical world, it raises the question of whether they can exhibit metaphorical tendencies akin to those of humans. Unlike their single-modality counterparts, MLLMs undergo alignment training with other modalities, granting them some degree of interaction with physical experiences. This leads us to wonder whether MLLMs are better equipped to understand semantic size and exhibit metaphorical associations that more closely resemble human cognition.

By examining whether LLMs demonstrate similar metaphorical associations, we aim to gain valuable insights into their cognitive processes. Specifically, testing how LLMs link abstract and concrete words based on semantic size allows us to assess whether their cognitive strategies mirror those of humans. If LLMs show comparable associations between semantic size and metaphorical reasoning, it could indicate that, in some ways, their cognitive processes align with human reasoning in the context of semantic size, offering insights into their potential for conceptual understanding.

\subsection{Dataset Construction}
\citet{yao2022can} employed a forced association task in which participants were presented with abstract concepts and required to choose one of three concrete word triplets to metaphorically complete a sentence (`Which metaphor do you find most fitting? To roar is like a(n)...') by selecting the most appropriate metaphorical association from the given concrete objects with either matching or varying semantic sizes.
However, the dataset consisted of only 110 questions, which may be insufficient in scale for testing LLMs effectively. To enhance the robustness and validity of our experimental results, we expanded the original dataset.

Examples from the final dataset are shown in Figure \ref{fig:dataset}, while more detailed examples in a multiple-choice question format can be found in Figure \ref{fig:dataset_large} in Appendix \ref{append:metaphor_dataset}. As illustrated, the dataset contains two types of triplets: one type includes size-varying triplets, where the semantic size labels of the three concrete words are big, medium, and small, respectively; the other type consists of size-matched triplets, where all three concrete words share similar semantic size labels. During the dataset construction process, we ensured that (1) the word frequency of both the abstract word pairs and concrete word triplets was comparable across the corpus and (2) the word length was identical across conditions. This helped eliminate any potential confounding effects of word frequency and length on the final results. The specific dataset construction methodology is outlined below:

\begin{figure}
    \centering
    \includegraphics[width=1\linewidth]{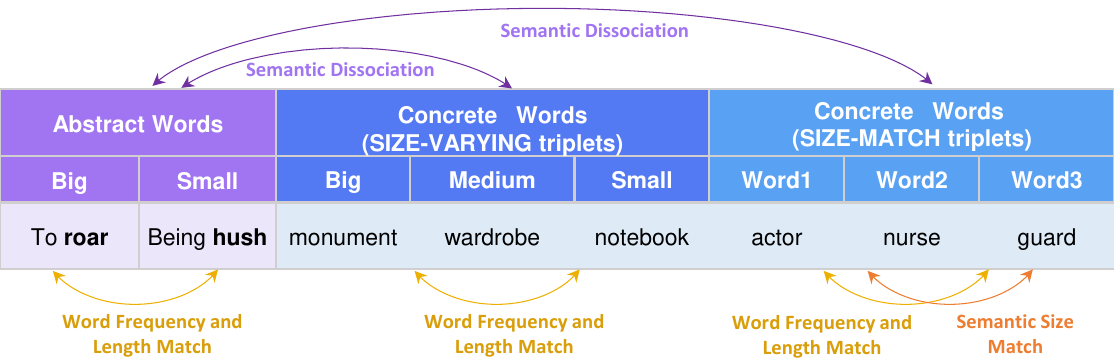}
    \caption{Examples for Semantic Size Metaphor dataset. Both abstract words pairs and concrete words triplets are matched across conditions for word frequency and length. SIZE-VARYING triplets: the semantic size labels of the three concrete words are big, medium, and small, respectively; SIZE-MATCH triplets: all three concrete words share similar semantic size.}
    \label{fig:dataset}
\end{figure}

\subsubsection{Semantic Size and Concreteness Classification}
\label{sec:glasgow}
We utilized the ratings from the Glasgow Norms \cite{scott2019glasgow}, a comprehensive psycholinguistic dataset, to classify words based on their semantic size. The Glasgow Norms provides normative ratings for 5,553 English words across nine dimensions: arousal, valence, dominance, concreteness, imageability, familiarity, age of acquisition, semantic size, and gender association. This dataset is particularly valuable for its large corpus, consistent participant ratings across dimensions, and the inclusion of novel dimensions such as semantic size and gender association.

In this study, we used the concreteness (CNC) dimension from the Glasgow Norms to classify words as either abstract or concrete. Words with a CNC score greater than 5 were categorized as concrete, while those with a score less than 3 were classified as abstract. Additionally, we used the semantic size dimension (SIZE, ranging from 1 to 7) to further divide words based on size: words with a score greater than 5 were considered `big', those between 3.5 and 4.5 were categorized as `middle', and those with a score below 3 were classified as `small'. A detailed description of the Glasgow Norms along with the definitions and labels for concreteness and semantic size can be found in Appendix Table \ref{tab:Glasgow}. Using this approach, we categorized the words into abstract (big, small) and concrete (big, middle, small) groups.

\subsubsection{Abstract Word Pairs and Concrete Word Triplets Construction}
After obtaining abstract and concrete words with different sizes, we needed to construct pairs of big and small abstract words, as well as triplets of concrete words with varying (or matched) sizes. To ensure that word frequency and length were not confounding factors, we matched the word pairs and triplets for frequency and length across conditions. The specific method is detailed below:

\textbf{a) Construction of Abstract Word Pairs:}  
In our expanded dataset, there were 128 abstract words. Half of the words described relatively big concepts (e.g., \textit{Freedom}), while the other half described relatively small concepts (e.g., \textit{Hush}). Words were matched for word length (number of letters) and word frequency on an item-by-item basis. 

For word length, both the abstract words and the concrete words (big, middle, and small) had the same number of letters, preventing either humans or LLMs from selecting based on word length.

For word frequency, we calculated the log occurrences per million for each word $i$ in the Brown Corpus \cite{francis1979brown} using the following formula:
$$\text{{log\_occur}}(i) = \log\left(\frac{{\text{{n}}_{\text{word}}^{i}}}{{\text{{N}}}} \times 10^6 \right),$$
where $n^i_{word}$ is the frequency of word $i$ and $N$ is the total number of words in the corpus.

This method provides a more balanced and robust way to handle word frequency data. We ensured that the difference in log occurrences between big and small abstract words, as well as between the big, middle, and small concrete words, was less than 2, thus ensuring that the words had comparable frequencies.

\textbf{b) Construction of Size-Varying Concrete Word Triplets:}  
The method for constructing size-varying concrete word triplets is similar to that used for abstract word pairs. After obtaining concrete words categorized as big, middle, and small, we matched these triplets based on word frequency and word length, following the same procedure used for abstract word pairs.

Additionally, we performed semantic dissociation to ensure that size-dependent metaphorical associations were not confounded by semantic associations. To achieve this, we used LLM2VEC \cite{behnamghader2024llm2vec}\footnote{We used the `McGill-NLP/LLM2Vec-Mistral-7B-Instruct-v2-mntp-supervised' model for this task. The threshold of 0.6 for semantic dissociation was determined based on the maximum cosine similarity from the original dataset in \cite{yao2022can}.} to obtain 4096-dimensional embeddings for each word. We then ensured that each concrete word had a cosine similarity of $\leq 0.6$ with their respective big and small abstract words.

\textbf{c) Construction of Size-Match Concrete Word Triplets:}  
The construction of size-match concrete word triplets followed a similar process as the size-varying triplets, with the key difference being that instead of selecting concrete words with varying semantic sizes (big, middle, small), we selected concrete words with matching semantic size labels. In addition, we ensured that the semantic size labels of the concrete words, based on the Glasgow Norms, differed by no more than 0.3.

By applying the above methods to expand the dataset, we ultimately obtained a dataset consisting of 128 abstract words. Each pair of abstract words was associated with four sets of concrete word triplets (two size-matched triplets and two size-varying triplets), resulting in a total of 512 multiple-choice questions (256 size-varying questions and 256 size-matched questions).




\subsection{Experiment}

In this study, we conducted both human and LLM evaluations. Participants, either human or LLMs, were asked to complete sentences like `Which metaphor do you find most fitting? To roar is like a(n)...' by selecting the most appropriate metaphorical association from the given concrete objects (detailed examples can be found in Figure \ref{fig:dataset_large}). This allowed us to measure the similarities and differences between human and LLM metaphorical tendencies. 

\textbf{Human Evaluation:}
For the human evaluation, we recruited 74 participants. All participants had normal or corrected-to-normal vision and had not been diagnosed with any learning or language disorders (e.g., dyslexia). The experiment lasted approximately 20 minutes. Detailed information of participants, including gender, age, and educational background, can be found in Appendix \ref{append:participant}.

\textbf{LLM Evaluation:}
For the LLM evaluation, we tested six different multi-modal large language models along with their corresponding text-only LLM backbones. These included LLMs from the GPT \cite{gpt4}, Yi \cite{ai2024yi}, Qwen \cite{qwen}, Mistral \cite{jiang2023mistral}, Vicuna \cite{vicuna}, and Llama \cite{llama3modelcard}.

In this experiment, we calculated the average probability of participants or LLMs selecting concrete words with different labels (big, middle, small). Additionally, to illustrate how certain participants or LLMs were in their choices, we defined a certainty metric. The certainty \textbf{C} is calculated as follows:
$$\textbf{C}=\textit{MAX}(P_{big},P_{middle},P_{small})-\textit{AVG}(P_{big},P_{middle},P_{small}),$$
where $P_{big}$, $P_{middle}$, and $P_{small}$ represent the probabilities of selecting big, middle, and small words (or word 1, word 2 and word 3 in the size-match scenario) across all questions, respectively. $\textit{MAX}()$ denotes the maximum value, and $\textit{AVG}()$ denotes the average value of the probabilities. The primary goal of our analysis is to evaluate whether the model selects the appropriate concrete word based on semantic size (e.g., associating a "large" abstract word with a "large" concrete word). \textbf{Certainty acts as a secondary measure to assess the strength of the model's preference for the correct choice. }

\begin{table}[]
\centering
\scalebox{0.85}{
\begin{tabular}{@{}llrrrrrrrr@{}}
\toprule
\multirow{2}{*}{Model}                    & \multicolumn{1}{c}{Abstract} & \multicolumn{4}{c}{Size-Varying Concrete Triplets}                                                                                                              & \multicolumn{4}{c}{Size-Match Concrete Triplets}                                                                                                           \\ \cmidrule(l){3-10} 
                                          & \multicolumn{1}{c}{Size}                       & \multicolumn{1}{c}{Big}            & \multicolumn{1}{c}{Middle} & \multicolumn{1}{c}{Small}          & \multicolumn{1}{c|}{\textbf{C}} & \multicolumn{1}{c}{0}              & \multicolumn{1}{c}{1}     & \multicolumn{1}{c}{2}              & \multicolumn{1}{c}{\textbf{C}} \\ \midrule
\multirow{2}{*}{Human}                    & Big                   & \multicolumn{1}{c}{\textbf{48.53}} & \multicolumn{1}{c}{26.38}  & \multicolumn{1}{c}{25.08}          & \multicolumn{1}{r|}{15.20}     & \multicolumn{1}{c}{\textbf{39.87}} & \multicolumn{1}{c}{30.38} & \multicolumn{1}{c}{29.75}          & 6.54                          \\
                                          & Small                 & \multicolumn{1}{c}{33.97}          & \multicolumn{1}{c}{25.40}  & \multicolumn{1}{c}{\textbf{40.63}} & \multicolumn{1}{r|}{7.30}      & \multicolumn{1}{c}{30.56}          & \multicolumn{1}{c}{\textbf{37.69}} & \multicolumn{1}{c}{31.75} & 4.36                          \\ \midrule
\multirow{2}{*}{GPT-3.5-turbo$^*$\cite{gpt4}}                             & Big                   & \textbf{49.53}                     & 26.09                      & 24.38                              & \multicolumn{1}{r|}{16.20}     & 35.63                              & \textbf{35.78}            & 28.59                              & 2.45                          \\
                                          & Small                 & 24.53                              & 26.72                      & \textbf{48.75}                     & \multicolumn{1}{r|}{15.42}     & 33.75                              & \textbf{36.09}            & 30.16                              & 2.76                          \\
\multirow{2}{*}{GPT-4-turbo \cite{gpt4}}              & Big                   & \textbf{52.66}                     & 27.19                      & 20.16                              & \multicolumn{1}{r|}{19.32}     & \textbf{35.94}                     & 32.03                     & 32.03                              & 2.60                          \\
                                          & Small                 & 31.56                              & \textbf{35.31}             & 33.13                              & \multicolumn{1}{r|}{1.98}      & 32.66                              & \textbf{32.81}            & 34.53                              & 1.20                          \\
\multirow{2}{*}{GPT-4o \cite{gpt4}}                   & Big                   & \textbf{50.47}                     & 28.28                      & 21.25                              & \multicolumn{1}{r|}{17.14}     & \textbf{39.84}                     & 32.66                     & 27.50                              & 6.51                          \\
                                          & Small                 & 26.41                              & 30.63                      & \textbf{42.97}                     & \multicolumn{1}{r|}{9.64}      & 27.81                              & \textbf{38.75}            & 33.44                              & 5.42                          \\
\multirow{2}{*}{GPT-4o-mini \cite{gpt4}}              & Big                   & \textbf{55.47}                     & 23.13                      & 21.41                              & \multicolumn{1}{r|}{22.14}     & \textbf{36.88}                     & 35.94                     & 27.19                              & 3.54                          \\
                                          & Small                 & 29.06                              & 25.63                      & \textbf{45.31}                     & \multicolumn{1}{r|}{11.98}     & \textbf{37.50}                     & 32.34                     & 30.16                              & 4.17                          \\ \midrule
\multirow{2}{*}{Yi-6B-Chat$^*$ \cite{ai2024yi}}               & Big                   & \textbf{39.45}                     & 27.73                      & 32.81                              & \multicolumn{1}{r|}{6.12}      & \textbf{35.39}                     & 35.00                     & 29.61                              & 2.06                          \\
                                          & Small                 & 34.38                              & 27.66                      & \textbf{37.97}                     & \multicolumn{1}{r|}{4.64}      & \textbf{36.17}                     & 31.80                     & 32.03                              & 2.84                          \\
\multirow{2}{*}{Yi-VL-6B \cite{ai2024yi}}                 & Big                   & \textbf{46.17}                     & 21.56                      & 32.27                              & \multicolumn{1}{r|}{12.84}     & 33.44                              & \textbf{34.06}            & 32.50                              & 0.73                          \\
                                          & Small                 & 30.00                              & 32.81                      & \textbf{37.19}                     & \multicolumn{1}{r|}{3.85}      & \textbf{34.45}                     & 33.75                     & 31.80                              & 1.12                          \\ \midrule
\multirow{2}{*}{Yi-34B-Chat$^*$ \cite{ai2024yi}}              & Big                   & \textbf{49.30}                     & 27.27                      & 23.44                              & \multicolumn{1}{r|}{15.96}     & 34.38                              & \textbf{35.31}            & 30.31                              & 1.98                          \\
                                          & Small                 & \textbf{35.94}                     & 28.13                      & \textbf{35.94}                     & \multicolumn{1}{r|}{2.60}      & \textbf{36.72}                     & 35.23                     & 28.05                              & 3.39                          \\
\multirow{2}{*}{Yi-VL-34B \cite{ai2024yi}}                & Big                   & \textbf{45.47}                     & 29.77                      & 24.77                              & \multicolumn{1}{r|}{12.14}     & 29.61                              & \textbf{36.72}            & 33.67                              & 3.39                          \\
                                          & Small                 & 26.80                              & 32.73                      & \textbf{40.47}                     & \multicolumn{1}{r|}{7.14}      & \textbf{36.09}                     & 35.78                     & 28.13                              & 2.76                          \\ \midrule
\multirow{2}{*}{Qwen-7B-Chat$^*$ \cite{qwen}}             & Big                   & \textbf{41.41}                     & 28.05                      & 30.55                              & \multicolumn{1}{r|}{8.07}      & 33.36                              & 31.80                     & \textbf{34.84}                     & 1.51                          \\
                                          & Small                 & \textbf{42.97}                     & 29.69                      & 27.34                              & \multicolumn{1}{r|}{9.64}      & 33.67                              & 31.56                     & \textbf{34.77}                     & 1.43                          \\
\multirow{2}{*}{Qwen-VL-Chat \cite{qwen}}             & Big                   & \textbf{47.66}                     & 27.50                      & 24.84                              & \multicolumn{1}{r|}{14.32}     & \textbf{34.45}                     & 33.36                     & 32.19                              & 1.12                          \\
                                          & Small                 & \textbf{45.39}                     & 27.66                      & 26.95                              & \multicolumn{1}{r|}{12.06}     & 32.89                              & \textbf{36.95}            & 30.16                              & 3.62                          \\
\multirow{2}{*}{Qwen-VL-max \cite{qwen}}              & Big                   & \textbf{35.63}                     & 29.61                      & 34.77                              & \multicolumn{1}{r|}{2.29}      & \textbf{37.11}                     & 33.20                     & 29.69                              & 3.78                          \\
                                          & Small                 & 20.16                              & 30.78                      & \textbf{49.06}                     & \multicolumn{1}{r|}{15.73}     & \textbf{35.47}                     & 33.05                     & 31.48                              & 2.14                          \\
\multirow{2}{*}{Qwen-VL-plus \cite{qwen}}             & Big                   & \textbf{34.53}                     & 33.59                      & 31.88                              & \multicolumn{1}{r|}{1.20}      & \textbf{37.34}                     & 36.02                     & 26.64                              & 4.01                          \\
                                          & Small                 & 24.92                              & 35.55                      & \textbf{39.53}                     & \multicolumn{1}{r|}{6.20}      & 33.83                              & \textbf{34.53}            & 31.64                              & 1.20                          \\ \midrule
\multirow{2}{*}{Mistral-7B-Instruct-v0.2$^*$ \cite{jiang2023mistral}} & Big                   & \textbf{45.23}                     & 28.67                      & 26.09                              & \multicolumn{1}{r|}{11.90}     & \textbf{37.58}                     & 32.19                     & 30.23                              & 4.24                          \\
                                          & Small                 & 31.25                              & 28.98                      & \textbf{39.77}                     & \multicolumn{1}{r|}{6.43}      & \textbf{37.19}                     & 32.58                     & 30.23                              & 3.85                          \\
\multirow{2}{*}{Llava-v1.6-Mistral-7b-hf \cite{llava}} & Big                   & \textbf{44.92}                     & 31.64                      & 23.44                              & \multicolumn{1}{r|}{11.59}     & \textbf{35.23}                     & 33.83                     & 30.94                              & 1.90                          \\
                                          & Small                 & 30.86                              & \textbf{35.86}                     & 33.28                     & \multicolumn{1}{r|}{2.53}      & \textbf{37.97}                     & 33.36                     & 28.67                              & 4.64                          \\ \midrule
\multirow{2}{*}{Vicuna-7b-v1.5$^*$ \cite{vicuna}}           & Big                   & 33.28                              & 32.73                      & \textbf{33.98}                     & \multicolumn{1}{r|}{0.65}      & 32.81                              & \textbf{35.94}            & 31.25                              & 2.60                          \\
                                          & Small                 & \textbf{34.61}                     & 34.38                      & 31.02                              & \multicolumn{1}{r|}{1.28}      & 35.00                              & \textbf{35.23}            & 29.77                              & 1.90                          \\
\multirow{2}{*}{Llava-v1.6-Vicuna-7b-hf \cite{llava}}  & Big                   & \textbf{37.50}                     & 28.91                      & 33.59                              & \multicolumn{1}{r|}{4.17}      & 34.69                              & \textbf{35.23}            & 30.08                              & 1.90                          \\
                                          & Small                 & 29.30                              & 32.34                      & \textbf{38.36}                     & \multicolumn{1}{r|}{5.03}      & 34.53                              & \textbf{35.00}            & 30.47                              & 1.67                          \\ \midrule
\multirow{2}{*}{Vicuna-13b-v1.5$^*$ \cite{vicuna}}          & Big                   & \textbf{37.19}                     & 30.39                      & 32.42                              & \multicolumn{1}{r|}{3.85}      & \textbf{33.59}                     & 33.28                     & 33.13                              & 0.26                          \\
                                          & Small                 & \textbf{35.08}                     & 31.02                      & 33.91                              & \multicolumn{1}{r|}{1.74}      & 31.95                              & \textbf{34.30}            & 33.75                              & 0.96                          \\
\multirow{2}{*}{Llava-v1.6-Vicuna-13b-hf \cite{llava}} & Big                   & \textbf{39.30}                     & 27.42                      & 33.28                              & \multicolumn{1}{r|}{5.96}      & 32.27                              & \textbf{34.53}            & 33.20                              & 1.20                          \\
                                          & Small                 & 29.69                              & 28.75                      & \textbf{41.56}                     & \multicolumn{1}{r|}{8.23}      & 32.19                              & \textbf{35.63}            & 32.19                              & 2.29                          \\ \midrule
\multirow{2}{*}{Meta-Llama-3-8B-Instruct$^*$ \cite{llama3modelcard}} & Big                   & 32.66                              & 32.42                      & \textbf{34.92}                     & \multicolumn{1}{r|}{1.59}      & 32.97                              & \textbf{33.59}            & 33.44                              & 0.26                          \\
                                          & Small                 & 31.72                              & 32.81                      & \textbf{35.47}                     & \multicolumn{1}{r|}{2.14}      & 33.05                              & 33.05                     & \textbf{33.91}                     & 0.57                          \\
\multirow{2}{*}{CogVLM2-llama3-Chat-19B \cite{hong2024cogvlm2}}  & Big                   & \textbf{35.70}                     & 31.48                      & 32.81                              & \multicolumn{1}{r|}{2.37}      & \textbf{36.41}                     & 32.19                     & 31.41                              & 3.07                          \\
                                          & Small                 & 30.16                              & 30.16                      & \textbf{39.69}                     & \multicolumn{1}{r|}{6.35}      & \textbf{36.72}                     & 32.27                     & 31.02                              & 3.39                          \\
\multirow{2}{*}{MiniCPM-Llama3-V-2\_5 \cite{yao2024minicpmvgpt4vlevelmllm}}    & Big                   & \textbf{49.22}                     & 25.23                      & 25.55                              & \multicolumn{1}{r|}{15.89}     & \textbf{41.09}                     & 33.44                     & 25.47                              & 7.76                          \\
                                          & Small                 & 22.81                              & 31.72                      & \textbf{45.47}                     & \multicolumn{1}{r|}{12.14}     & 32.97                              & \textbf{36.56}            & 30.47                              & 3.23                          \\ \bottomrule
\end{tabular}
}
\caption{Results for Semantic Size Metaphor Study using the extended dataset.* denotes text-only LLM models, and \textbf{C} represents the certainty metric. Human participants consistently demonstrated a tendency to select big (small) concrete words for big (small) abstract words. All experimental results represent the average values across 10 runs.}
\label{tab:extend}
\end{table}

\subsection{Results}

\begin{figure}
    \centering
    \includegraphics[width=1\linewidth]{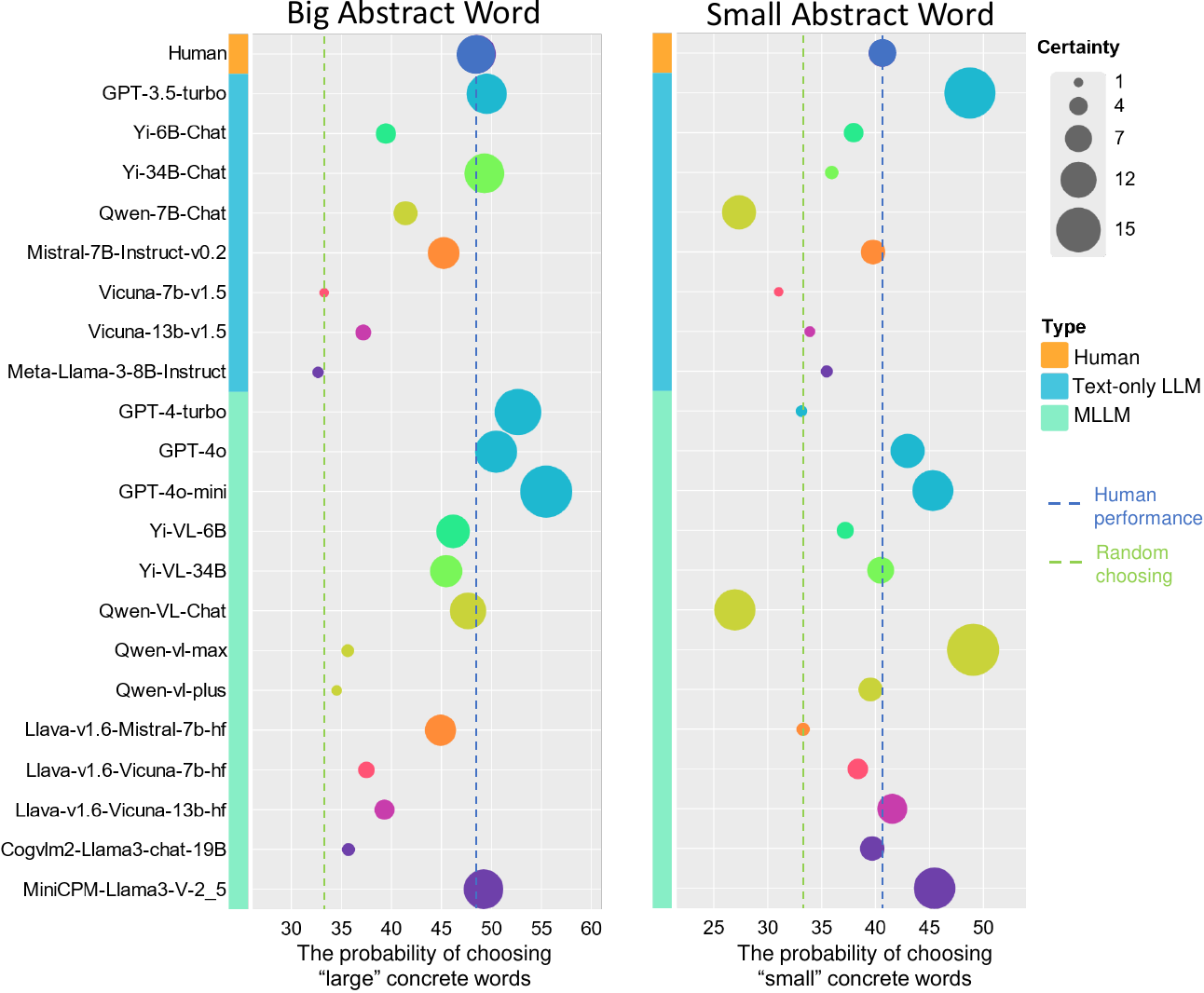}
    \caption{Results for Semantic Size Metaphor Study with size-varying setting using the extended dataset. The vertical axis represents different models, with color distinguishing the type: human (orange), text-only LLM (blue), and multi-modal LLM (MLLM, green). The horizontal axis shows the probability of selecting the corresponding large or small concrete object for a given abstract word. Each bubble’s color represents models from the same family, while the bubble size indicates the model's certainty in making a selection, with larger bubbles representing greater certainty. An LLM that performs similarly to humans should have a higher probability and when presented with big or small abstract words, reflecting accuracy and confidence in selecting size-congruent concrete objects. As shown in the figure, MLLMs tend to have larger and further-right bubbles compared to text-only LLMs, indicating improved performance after multi-modal training.}
    \label{fig:exp1_result}
\end{figure}

The results of the Semantic Size Metaphor study are shown in Table \ref{tab:extend}. To present the findings more intuitively, we visualize the experimental results, as shown in Figure \ref{fig:exp1_result}. Based on the results, we summarize the following key findings:

\textbf{1) Multi-modal training enhances semantic size understanding:}
From the table, we can observe that (M)LLMs demonstrate some understanding of semantic size. However, multi-modal trained MLLMs exhibit tendencies that align more closely with human participants compared to their text-only counterparts.
In other words, in the size-varying experimental setup, MLLMs are more likely to associate large abstract words with large concrete words and small abstract words with small concrete words. In the size-match setup, they tend to select more randomly, and all models show lower certainty in size-match conditions compared to size-varying conditions.
For example, in the `Size-Varying Triplets' task, MLLMs like GPT-4o and Qwen-VL-max, compared to GPT-3.5 and Qwen-7B, tend to select big (or small) concrete words for big (or small) abstract words with a higher certainty metric (\textbf{C}), which measures the confidence of the model in choosing the most metaphorically appropriate word. In most cases, MLLMs have a higher certainty metric across multiple conditions compared to LLMs, indicating greater confidence in their selections.

A more intuitive example is the Vicuna and LLaMA-3-8B models. In the text-only versions of Vicuna (both 13B and 7B) and LLaMA-3-8B-Instruct, the models failed to exhibit human-like metaphorical tendencies. The Vicuna 7B model, in particular, demonstrated an inverse tendency compared to humans, with very low certainty, indicating a poor understanding of the semantic size relationship between abstract and concrete words. However, interestingly, after multi-modal training with Llava, as well as training with CogVLM and MiniCPM, both Vicuna 13B and 7B, as well as LLaMA-3-8B, exhibited human-like metaphorical associations and significantly higher certainty.

Importantly, the size congruency effects observed in this experiment cannot be explained by differences in word frequency or length between the candidate words, nor by differences in semantic associations between the target word and the three concrete objects. The careful control of these variables ensures that the metaphorical associations observed reflect the model's semantic size reasoning processes rather than statistical or semantic associations biases.

These results support our hypothesis that \textbf{abstract concepts can be metaphorically associated with concrete objects based on their size, and that multi-modal training improves the models' ability to align more closely with human cognition}. The fact that MLLMs generally exhibit more human-like tendencies and higher certainty suggests that incorporating multiple modalities (e.g., text and visual data) helps these models develop a more nuanced and human-like understanding of abstract and concrete associations, particularly regarding semantic size.

\textbf{2) Larger models exhibit better semantic size understanding:}
For instance, when comparing the Yi-34B and Yi-6B models or the Vicuna-7B and Vicuna-13B models, the larger versions consistently achieve higher certainty scores, indicating a more refined understanding of semantic size.

\textbf{3) Better multi-modal training leads to improved semantic size comprehension:}
For example, Qwen-VL-max outperforms the more basic multi-modal models, such as Qwen-VL-Chat and Qwen-VL-plus. As shown in the table, Qwen-VL-Chat even performs worse than Qwen-7B-Chat. In contrast, more advanced multi-modal training, like that used in Qwen-VL-plus and Qwen-VL-max\footnote{https://github.com/QwenLM/Qwen-VL}, brings notable benefits to text-only LLMs. 

In comparison to Qwen-VL-plus, according to their technical report \cite{bai2023qwen_vl}, Qwen-VL-max demonstrates further improvements in visual reasoning and instruction-following capabilities, offering a more advanced level of visual perception and cognitive understanding. This model delivers optimal performance across a broader range of complex tasks. This indicates that insufficient multi-modal training can negatively affect an LLM's ability to comprehend semantic size. Similar patterns can be observed in the comparisons between GPT-4 and GPT-4o, as well as between CogVLM and MiniCPM. \textbf{Stronger MLLMs generally show human-like choices with higher certainty, reinforcing the importance of advanced multi-modal training in enhancing model cognition}.

\subsection{Further Exploration}
We also conducted experiments using the original, unexpanded dataset constructed by \cite{yao2022can}, with the detailed results provided in Appendix \ref{append:semantic size prompt}.

To further investigate whether LLMs can make metaphorical associations based on semantic size, we simplified the task by directly prompting the models with, `Consider the semantic size of the given abstract word.' The specific input format and corresponding experimental results can be found in Appendix \ref{append:semantic size prompt}.  

As shown in Table  \ref{tab:exp1_extend_prompt_size}, most models exhibited a closer alignment with human reasoning when explicitly guided by the semantic size prompt. However, certain models, such as Yi-VL-6B and Qwen-VL-Chat, continued to face challenges in understanding semantic size, despite receiving direct instructions.

This limitation in metaphorical reasoning led us to explore whether LLMs can effectively encode semantic size in their internal representations, beyond the explicit prompting provided. Thus, in the second exploration, we shift our focus to probing the internal representations of LLMs, assessing whether semantic size is inherently captured by these models. 

\section{Internal Exploration: Probing How LLMs Encode Semantic Size.}
\label{sec:in}

In this study, we extend our exploration of semantic size understanding by examining the internal representations of LLMs. Rather than solely analyzing the generated text, we aim to determine whether LLMs encode the semantic size of different words in their internal activations. This allows us to investigate the underlying structures that support their ability to comprehend semantic size.

\subsection{Method}

We first use the probing techniques to investigate the internal representations of LLMs. Probing refers to the use of linear classifiers trained independently of the model to predict a specific attribute \cite{DBLP:conf/iclr/AlainB17, DBLP:journals/coling/Belinkov22}, in this case, the semantic size (big or small, as defined in the previous study) of a given word. By applying semantic size probing, we aim to better understand the roles and dynamics of the intermediate layers of the model.

Specifically, we first prompt the model with a target word and capture the attention head activations at the final token position, denoted as $\mathbf{X} \in \mathbb{R}^{L \times H \times D}$, where $L$ is the number of layers, $H$ is the number of attention heads per layer, and $D$ is the activation dimension. We then train individual probes for each attention head at every layer to classify the semantic size labels. Given word $i$, for a specific activation $x^l_h(i)\in \mathbb{R}^{D}$ from layer $l$ and head $h$, we employ a logistic regression model to predict the probability of the word's semantic size being large:
$$\hat{y}^l_h(i) = \sigma(\mathbf{W}x^l_h(i) + b),$$
where $\sigma(·)$ is the logistic sigmoid function, $\mathbf{W} \in \mathbb{R}^{D}$ is the weight vector, and $b \in \mathbb{R}$ is the bias term. The parameters $\mathbf{W}$ and $b$ are optimized by minimizing the cross-entropy loss function:
$$\mathcal{L}^l_h(W, b) = -\frac{1}{N} \sum _{i=1} ^ N \left( y^l_h(i) \log(\hat{y}^l_h(i)) + (1 - y^l_h(i)) \log(1 - \hat{y}^l_h(i)) \right),$$
where $N$ is the dataset size. Probes are independently trained for each attention head at every layer. This approach provides deeper insights into the internal mechanisms of LLMs with respect to semantic size comprehension.

For this experiment, we use the abstract and concrete words from the dataset constructed in the previous exploration. The dataset consists of 128 abstract words (64 large and 64 small) and 256 concrete words (128 large and 128 small). The dataset was split into training and testing sets with a 7:3 ratio.

\subsection{Results}

\begin{table}[]
\centering
\begin{tabular}{@{}lcccc@{}}
\toprule
                         & \multicolumn{2}{c}{Abstract}                                      & \multicolumn{2}{c}{Concrete}                 \\ \cmidrule(l){2-5} 
Model                    & Val\_Acc       & \multicolumn{1}{c|}{Roc\_Auc}       & Val\_Acc       & Roc\_Auc       \\ \midrule
Mistral-7B-Instruct-v0.2$^*$ \cite{jiang2023mistral} & 48.50          & \multicolumn{1}{c|}{70.40}          & 51.39          & 75.78          \\
Llava-v1.6-Mistral-7b-hf \cite{llava}& \textbf{49.35} & \multicolumn{1}{c|}{\textbf{73.74}} & \textbf{53.50} & \textbf{80.93} \\ \midrule
Vicuna-7b-v1.5$^*$  \cite{vicuna}         & 57.92          & \multicolumn{1}{c|}{66.81}          & 67.18          & 78.24          \\
Llava-v1.6-Vicuna-7b-hf \cite{llava} & \textbf{58.14} & \multicolumn{1}{c|}{\textbf{67.46}} & \textbf{67.60} & \textbf{79.16} \\ \midrule
Vicuna-13b-v1.5$^*$   \cite{vicuna}        & 57.30          & \multicolumn{1}{c|}{66.27}          & 67.34          & 77.55          \\
Llava-v1.6-Vicuna-13b-hf \cite{llava} & \textbf{57.98} & \multicolumn{1}{c|}{\textbf{67.45}} & \textbf{69.49} & \textbf{79.66} \\ \midrule
Yi-6B-Chat$^*$    \cite{ai2024yi}         & 54.39          & \multicolumn{1}{c|}{\textbf{65.85}} & 62.13          & 78.02          \\
Yi-VL-6B        \cite{ai2024yi}       & \textbf{57.39} & \multicolumn{1}{c|}{65.73}          & \textbf{68.12} & \textbf{79.50} \\ \midrule
Yi-34B-Chat$^*$    \cite{ai2024yi}      & 59.91          & \multicolumn{1}{c|}{\textbf{69.83}} & 68.10          & \textbf{82.46} \\
Yi-VL-34B       \cite{ai2024yi}    & \textbf{60.87} & \multicolumn{1}{c|}{66.97}          & \textbf{69.58} & 77.15          \\ \midrule
Meta-Llama-3-8B-Instruct$^*$ \cite{llama3modelcard} & 50.75          & \multicolumn{1}{c|}{\textbf{71.97}} & 58.90          & \textbf{84.80} \\
CogVLM2-llama3-Chat-19B \cite{hong2024cogvlm2} & \textbf{57.99} & \multicolumn{1}{c|}{69.09}          & \textbf{64.48} & 76.51          \\ \midrule
Qwen-7B-Chat$^*$     \cite{qwen}        & \textbf{64.12} & \multicolumn{1}{c|}{\textbf{70.86}} & \textbf{75.20} & \textbf{84.01} \\
Qwen-VL-Chat      \cite{qwen}    & 61.51          & \multicolumn{1}{c|}{68.11}          & 72.53          & 81.59          \\ \bottomrule
\end{tabular}
\caption{Results for Semantic Size Probing. $^*$ denotes text-only LLM models and Val\_Acc and Roc\_Auc denotes evaluation accuracy and Roc\_Auc Score \cite{hand2001simple}. All experimental results represent the average values across 10 runs.}
\label{tab:exp2_result}
\end{table}

\begin{figure}
    \centering
    \includegraphics[width=1\linewidth]{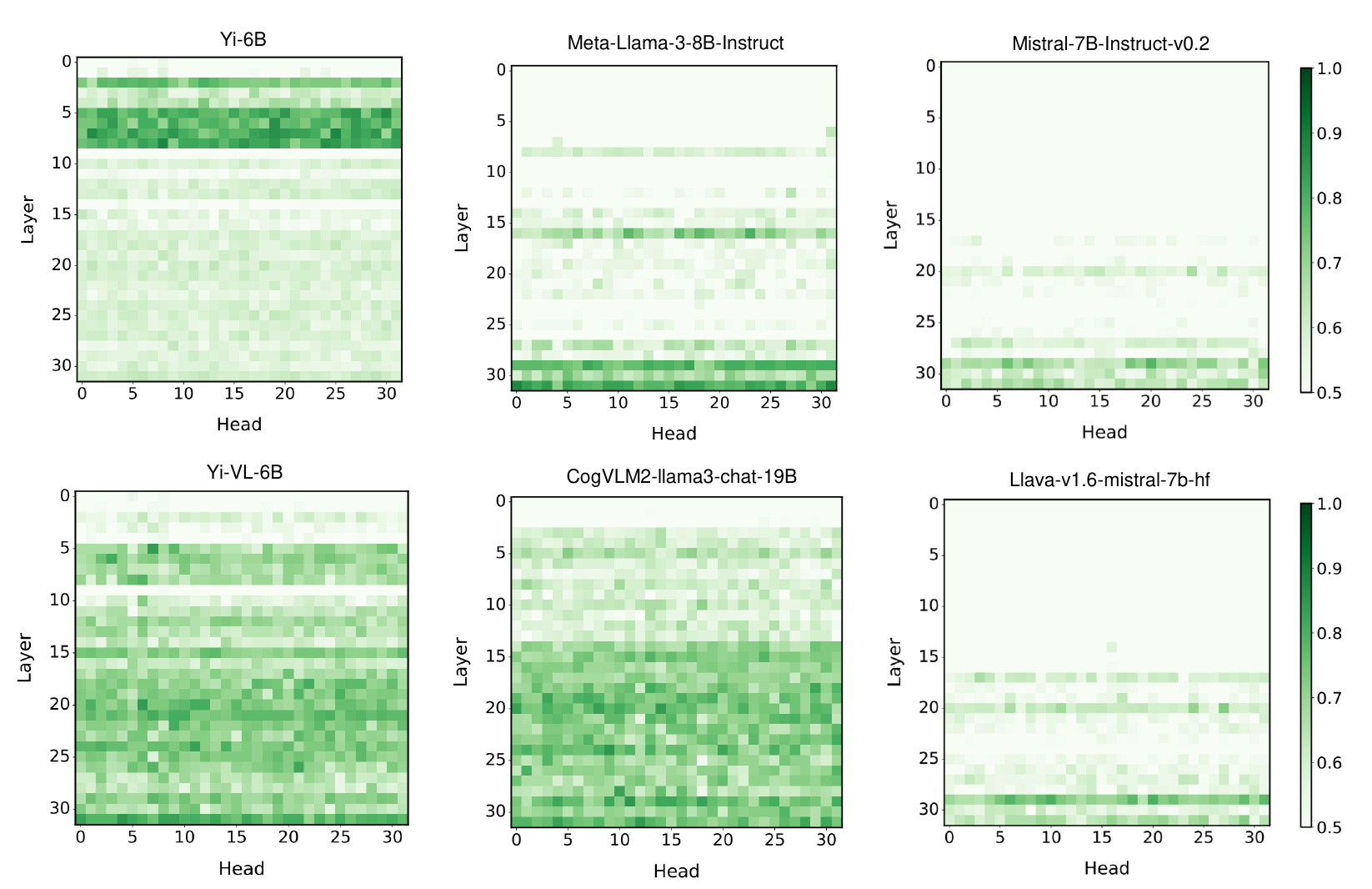}
    \caption{Semantic size probing accuracy (average of abstract and concrete words) based on attention head activations across all layers. The multi-modal LLMs (bottom row) show deeper shading, indicating higher accuracy compared to the text-only LLMs (top row). Notably, the Mistral model's accuracy remains low, even though multi-modal training improves performance slightly. This aligns with the findings from the previous study 1, where Mistral exhibited lower overall effectiveness in semantic size metaphor.}
    \label{fig:exp2_probe_res}
\end{figure}

The experimental results \footnote{We conducted probing experiments on the open-sourced models from the previous study 1. However, for models that have not been open-sourced, such as ChatGPT, Qwen-VL-Plus, and Qwen-VL-Max, we were unable to access their internal representations for probing. } are presented in Table \ref{tab:exp2_result}. We also visualize the accuracy of the probes in Figure \ref{fig:exp2_probe_res}. From the table and the figure, we can derive two key conclusions:

\textbf{1) Multi-modal LLMs are more effective at classifying the semantic size of words.} 
From the table, we can see that most (M)LLMs' attention can, to some extent, encode semantic size. Multi-modal training further enhances LLMs' ability to understand the semantic size of both abstract and concrete words, with MLLMs generally outperforming their text-only counterparts. 
This aligns with findings in psychology \cite{yao2022can}, which suggest that \textit{`The semantic size of abstract concepts can indeed be represented in physical (visual) size, as a function of context (font size variation) and task demands (forced associations, lexical decision, size vs emotion judgements).'} Through multi-modal training, MLLMs develop stronger associations between text and images, leading to a better understanding of concrete, physical words. This enhanced ability to represent concrete concepts, in turn, allows MLLMs to more effectively interpret and represent abstract words.

However, a few exceptions were observed. For instance, both the text-only and multi-modal Mistral models exhibited classification accuracies below 50\%, indicating difficulties in grasping semantic size. This observation is consistent with the results from Study 1, where the multi-modal Llava-v1.6-Mistral-7B-hf often associated small abstract words with medium-sized concrete words. Similarly, Qwen-VL-Chat demonstrated lower performance in semantic size classification compared to its text-only counterpart, Qwen-7B-Chat, aligning with earlier external exploration findings where Qwen-VL-Chat struggled with semantic size metaphor.

While we could not perform probing experiments on more advanced models like Qwen-VL-plus and Qwen-VL-max due to the unavailability of their internal representations, the current findings further underscore the critical role of sufficient multi-modal training in enhancing a model's ability to grasp semantic size. This highlights the necessity of robust multi-modal training to achieve a more comprehensive understanding of abstract and concrete concepts.

\textbf{2) The semantic size of concrete words is more easily distinguished by LLMs. }As shown in the Table \ref{tab:exp2_result}, both multi-modal and text-only models demonstrate higher classification accuracies when categorizing the semantic size of concrete words. This suggests that LLMs have an easier time understanding the physical dimensions of concrete words and MLLMs showing even greater classification accuracy of concrete words. This further reinforces the idea that while LLMs may struggle with the abstract conceptualization of size, particularly in words with no physical referent, they excel in concrete word classification, particularly when trained on multiple modalities.

In summary, the results from the Study 2 emphasize the significance of multi-modal training in enhancing LLMs’ ability to grasp both abstract and concrete concepts, especially in terms of semantic size. The findings also suggest that while text-only LLMs are proficient at understanding the physical size of concrete words, the addition of multi-modal training can substantially improve their performance in this task.

\section{Real-world Exploration: Investigating the Impact of Semantic Size in Attention-grabbing Headlines.}
\label{sec:real}

As we move from the external semantic size metaphor study to the more internal semantic size probing study, we have demonstrated that LLMs possess a certain level of understanding of semantic size. Although not all LLMs exhibit this capability, and their current performance still lags behind that of humans, multi-modal training can efficiently enhances their ability to comprehend semantic size. This naturally leads us to consider whether these findings can be applied in real-world scenarios. 

To bridge this theoretical understanding with practical applications, it is valuable to investigate how LLMs deal with attention-grabbing contents, such as advertising or e-commerce, which typically involve larger semantic sizes.
Previous research has shown that semantic size and emotional arousal are positively correlated \cite{yao2013semantic}, and it is well-established that attention-grabbing headlines tend to evoke higher levels of emotional arousal, and thus have a larger semantic size compared to neutral headlines \cite{chen2015misleading, pengnate2019shocking}.

\begin{figure}[h]
    \centering
    \includegraphics[width=1\linewidth]{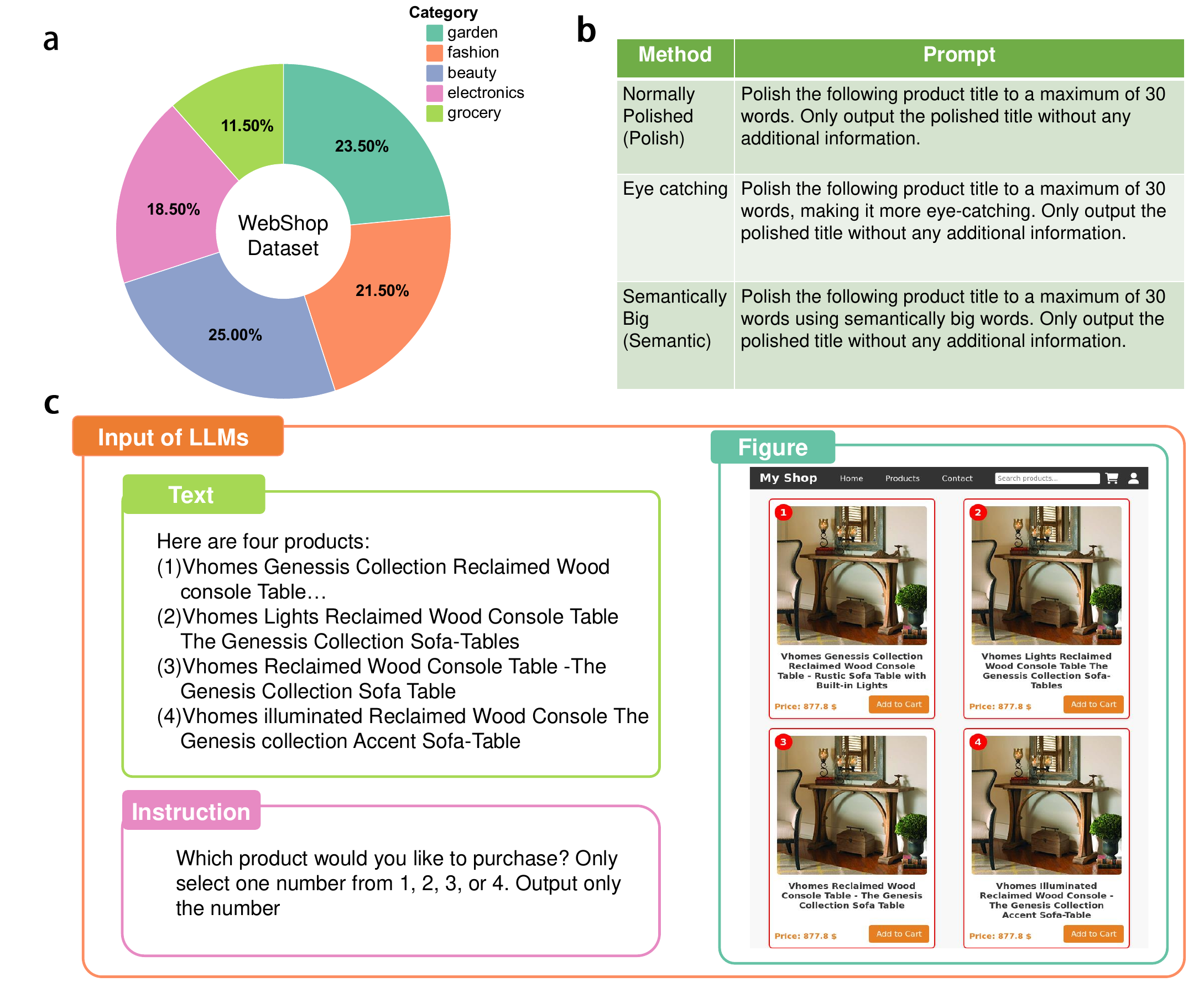}
    \caption{Overview of the dataset and experimental setup for Study 3: (a) Proportion of product categories selected from the WebShop dataset, including garden, fashion, beauty, electronics, and grocery. (b) The three polishing methods (Polish, Eye-catching, Semantic) and their corresponding prompts used to modify product titles. (c) Example of the web shopping page interface, where four variants of the same product, each with a different title (one unpolished and three polished using different methods), are displayed for comparison. All other elements on the page are controlled to ensure only the product titles vary.}
    \label{fig:exp3_dataset_construction}
\end{figure}

The rapid advancement of LLMs has also fueled the development of LLM-powered agents capable of following language instructions and executing actions in real-world or simulated environments \cite{DBLP:journals/corr/abs-2311-11797}. These agents are now proficient in tasks such as graphical user interface (GUI) automation \cite{DBLP:conf/icml/ZhengGK0024, ma2024coco}, and it is foreseeable that in the near future, LLM agents could be used in real-world scenarios such as online shopping. In such contexts, LLM agents may be responsible for navigating vast amounts of product information and advertisements, where the influence of attention-grabbing headlines and marketing strategies will become a relevant factor \cite{ma2024caution}. Therefore, understanding their biases toward attention-grabbing headlines becomes crucial. Building on this, we aim to explore whether LLMs exhibit a bias toward more eye-catching headlines with larger semantic sizes. Specifically, we pose the question: \textbf{`Are LLMs more likely to be influenced by attention-grabbing headlines with a larger semantic size?'}



\subsection{Dataset Construction}
In this study, we select the first 200 products from the WebShop dataset \cite{DBLP:conf/nips/Yao0YN22}, which contains over one million products scraped from Amazon.com. These products span various categories, including garden, fashion, beauty, electronics, and grocery, with the proportion of each category shown in Figure \ref{fig:exp3_dataset_construction}(a).

For each product title, we apply three different polishing methods: Normally Polished (Polish), Eye-catching (Eye\_catching), and Semantically Big (Semantic). Specifically, we used the prompts shown in Figure \ref{fig:exp3_dataset_construction}(b) to prompt GPT-4o to polish the original product titles. We also use the Glasgow Norms to calculate the average semantic size of product titles generated by each polishing method, as shown in Table \ref{tab:semantic_size_polish}. The Semantic polishing method yielded the highest semantic size, while the Original titles had the smallest average semantic size. Both the Polish and Eye-catching methods also increased the semantic size of the titles, though to a lesser extent than the Semantic method. 

Following this, we construct a web shopping interface, as illustrated in Figure \ref{fig:exp3_dataset_construction} (c). Each product had its own dedicated page, where four variants of the same product were displayed—one with the original, unpolished title, and three with titles polished using the aforementioned methods. We controlled all other variables on the page, ensuring that only the product titles differed across the variants.
\begin{table}[]
\centering
\begin{tabular}{@{}ccccc@{}}
\toprule
              & Original & Polish & Eye\_catching & Semantic \\ \midrule
Semantic Size & 3.83     & 3.90   & 3.99          & 4.05     \\ \bottomrule
\end{tabular}
\caption{Average semantic size of different polishing methods}
\label{tab:semantic_size_polish}
\end{table}

\subsection{Results}
In the experiment, we employed three different experimental setups: text-only, figure-only, and figure-text. Specifically, for the text-only condition, we provided the LLMs with only the text and instruction from Figure \ref{fig:exp3_dataset_construction}(c). For the figure-only condition, we provided the figure and instruction. Finally, in the figure-text condition, we supplied the text, figure, and instruction. The experimental results are presented in Table \ref{tab:exp3_results}.

\begin{table}[]
\centering
\begin{tabular}{@{}llcccc@{}}
\toprule
                              &                          & semantic       & eye\_catch     & polish         & original \\ \midrule
\multirow{19}{*}{Text only}   & GPT-3.5-turbo \cite{gpt4}           & \textbf{33.70} & 28.90          & 18.00          & 19.40    \\
                              & GPT-4o-mini    \cite{gpt4}          & \textbf{48.20} & 32.90          & 12.90          & 6.00     \\
                              & GPT-4o        \cite{gpt4}           & \textbf{47.80} & 27.60          & 16.70          & 7.90     \\ \cmidrule(l){2-6} 
                              & Vicuna-7b-v1.5  \cite{vicuna}         & 24.90          & 26.30          & \textbf{26.70} & 22.10    \\
                              & Llava-v1.6-Vicuna-7b-hf \cite{llava} & \textbf{34.75} & 23.00          & 16.05          & 26.20    \\ \cmidrule(l){2-6} 
                              & Vicuna-13b-v1.5   \cite{vicuna}   & \textbf{38.25} & 23.55          & 21.00          & 17.20    \\
                              & Llava-v1.6-Vicuna-13b-hf \cite{llava} & \textbf{32.50} & 24.95          & 24.70          & 17.85    \\ \cmidrule(l){2-6} 
                              & Mistral-7B-Instruct-v0.2 \cite{jiang2023mistral}& 23.60          & 26.85          & \textbf{27.80} & 21.75    \\
                              & Llava-v1.6-Mistral-7b-hf \cite{llava}& 23.25          & 27.05          & \textbf{27.60} & 22.10    \\ \cmidrule(l){2-6} 
                              & Yi-6B-Chat        \cite{ai2024yi}       & \textbf{44.95} & 23.85          & 16.30          & 14.90    \\
                              & Yi-VL-6B      \cite{ai2024yi}  & \textbf{36.80} & 26.90          & 22.60          & 13.70    \\ \cmidrule(l){2-6} 
                              & Yi-34B-Chat     \cite{ai2024yi}   & \textbf{31.55} & 28.70          & 21.95          & 17.80    \\
                              & Yi-VL-34B     \cite{ai2024yi}     & \textbf{50.40} & 25.85          & 11.55          & 12.20    \\ \cmidrule(l){2-6} 
                              & Meta-Llama-3-8B-Instruct \cite{llama3modelcard}& \textbf{36.30} & 24.45          & 21.40          & 17.85    \\
                              & CogVLM2-llama3-Chat-19B \cite{hong2024cogvlm2} & 21.90          & 25.65          & \textbf{29.35} & 23.10    \\
                              & MiniCPM-Llama3-V-2\_5  \cite{yao2024minicpm}  & \textbf{52.45} & 24.50          & 9.60           & 13.45    \\ \cmidrule(l){2-6} 
                            & Qwen-7B-Chat     \cite{qwen}     & 25.70          & 29.25          & \textbf{30.55} & 14.50    \\
                              & Qwen-VL-Chat       \cite{qwen}      & 23.45          & \textbf{27.20} & 26.30          & 23.05    \\

                              \midrule
\multirow{10}{*}{Figure+Text} & GPT-4o-mini    \cite{gpt4}       & \textbf{50.00} & 29.10          & 13.40          & 7.50     \\
                              & GPT-4o       \cite{gpt4}       & \textbf{30.50} & 29.60          & 27.70          & 12.20    \\
                              & Llava-v1.6-Vicuna-7b-hf \cite{llava} & \textbf{33.35} & 25.55          & 23.95          & 17.15    \\
                              & Llava-v1.6-Vicuna-13b-hf \cite{llava}& \textbf{40.40} & 27.80          & 18.15          & 13.65    \\
                              & Llava-v1.6-Mistral-7b-hf \cite{llava}& 23.35          & \textbf{27.55} & 27.35          & 21.75    \\
                              & Yi-VL-6B      \cite{ai2024yi}      & \textbf{33.75} & 27.80          & 19.40          & 19.05    \\
                              & Yi-VL-34B        \cite{ai2024yi}         & \textbf{35.20} & 27.35          & 20.20          & 17.25    \\
                              & CogVLM2-llama3-Chat-19B  \cite{hong2024cogvlm2}& \textbf{27.95} & 25.25          & 21.55          & 25.25    \\
                              & MiniCPM-Llama3-V-2\_5   \cite{yao2024minicpm} & \textbf{28.35} & 24.55          & 25.05          & 22.05    \\
                              & Qwen-VL-Chat    \cite{qwen}     & 27.45          & \textbf{31.60} & 18.80          & 22.15    \\ \midrule
\multirow{10}{*}{Figure only} & GPT-4o-mini     \cite{gpt4}      & \textbf{30.70} & 26.30          & 22.70          & 20.30    \\
                              & GPT-4o        \cite{gpt4}      & 25.25          & 26.25          & \textbf{27.50} & 21.00    \\
                              & Llava-v1.6-Vicuna-7b-hf \cite{llava}& 21.75          & 24.55          & \textbf{29.00} & 24.70    \\
                              & Llava-v1.6-Vicuna-13b-hf \cite{llava}& 25.15          & \textbf{26.75} & 26.65          & 21.45    \\
                              & Llava-v1.6-Mistral-7b-hf \cite{llava}& 21.80          & 26.75          & \textbf{28.90} & 22.55    \\
                              & Yi-VL-6B       \cite{ai2024yi}           & \textbf{26.75} & 24.15          & 24.45          & 24.65    \\
                              & Yi-VL-34B       \cite{ai2024yi}          & 22.40          & \textbf{27.15} & 26.75          & 23.70    \\
                              & CogVLM2-llama3-Chat-19B  \cite{hong2024cogvlm2}& \textbf{27.50} & 25.00          & 21.00          & 26.50    \\
                              & MiniCPM-Llama3-V-2\_5   \cite{yao2024minicpm} & \textbf{26.80} & 24.25          & 23.90          & 25.05    \\
                              & Qwen-VL-Chat    \cite{qwen}   & \textbf{30.00} & 23.50          & 21.50          & 25.00    \\ \bottomrule
\end{tabular}
\caption{Results for Real-world Exploration: Semantic Size and Attention-grabbing Headlines. Text-only: only text and instruction are provided. Figure-only: only figure and instruction are provided. Figure + Text:  text, figure, and instruction are all provided. All experimental results represent the average values across 10 runs.}
\label{tab:exp3_results}
\end{table}

The results presented in Table \ref{tab:exp3_results} reveal several important findings across different models and experimental settings:

\textbf{(1) LLMs are more attracted to titles with larger semantic size.}
For most LLMs, titles with larger semantic sizes (using the \textit{Semantic} polish method) are generally more appealing, especially compared to other polishing methods like eye-catching and polish. This trend is even more pronounced for MLLMs than for text-only LLMs, indicating that multi-modal training enhances models' sensitivity to semantic size. This finding aligns with our earlier studies, where models such as Mistral and Qwen, which demonstrated a weaker understanding of semantic size, tended to prefer titles polished using the polish method rather than the semantic method. This again emphasizes their limitations in comprehending semantic size, reinforcing the conclusions from the previous two studies.

\textbf{(2) MLLMs are more sensitive to text than visual modalities.}
A notable observation is that MLLMs exhibit a stronger sensitivity to the semantic size of words in the textual modality, especially when comparing the figure-only and text-only conditions. In the text-only condition, most models show a tendency for product titles with larger semantic size, reinforcing the notion that LLMs are more attuned to textual semantics than to visual inputs. However, when both image and text are presented together (figure$+$text condition), almost all MLLMs consistently chose titles with larger semantic size, further indicating that multi-modal training enhances their ability to interpret semantic size and subsequently makes them more emotionally involved in decision-making.

There are a few notable exceptions, such as Llava-v1.6-Mistral-7B-HF and Qwen-VL-Chat, which favored titles polished using the eye-catching method rather than those with larger semantic size. This is consistent with previous experimental results, where these models struggled to fully grasp the concept of semantic size. 

\textbf{(3) Emotional engagement and decision-making.}
The results also indicate that models trained with multi-modal inputs are not only better at understanding semantic size but are also more emotionally engaged in decision-making when presented with semantically larger titles. 
The tendency for most MLLMs to choose titles with larger semantic sizes when both image and text are available suggests that these models may rely on semantic size as a cue for importance or relevance in emotionally charged tasks, such as decision-making. This supports the idea that multi-modal training plays a crucial role in shaping how LLMs process emotional and semantic cues in real-world applications, such as advertising or product selection.

\section{Key Takeaways}
\paragraph{LLMs find `Spaceship' easier to grasp than `Love'}
Our experimental results indicate that the semantic size of abstract words poses a greater challenge for LLMs to understand. While multi-modal large language models (MLLMs) outperform their text-only counterparts, both categories of models still encounter difficulties in differentiating abstract words from concrete ones. Interestingly, humans face similar challenges in comprehending abstract words \cite{schwanenflugel2013abstract}. Like humans, LLMs tend to classify concrete words more readily in terms of size, while abstract terms such as `love' or `hush' present more complex hurdles. This suggests that although LLMs have made significant strides in understanding physical dimensions, their grasp of abstract semantic associations remains intricate, reflecting patterns of human-like cognition.

\paragraph{Multi-modal training bridges the gap for LLMs to understand `Love' like humans}
For humans, embodied experiences contribute significantly to the semantic size of words. In humans' understanding, abstract terms possess a sense of heft or breadth, often expressed through greater visual size in concrete language and abstract words may `borrow' this similar expressions of embodiment \cite{yao2022can}. Our studies reveal that, much like humans, multi-modal training enables large language models to relate to visual experiences. This enhancement allows LLMs not only to comprehend concrete words more effectively but also to gain a deeper understanding of abstract concepts. Across all studies, our results demonstrate that multi-modal large language models outperform text-only LLMs in leveraging semantic size. MLLMs align more closely with human cognitive processes, particularly in metaphorical reasoning and in selecting product titles that reflect larger semantic sizes. This underscores the critical importance of integrating multiple modalities in training, bringing LLMs one step closer to understanding abstract words like `love' as humans do.
 
\paragraph{LLMs can also fall into the clickbait trap and future implications}
Our Semantic Size and Attention-grabbing Headlines Study demonstrates that both MLLMs and text-only LLMs exhibit a bias toward titles with larger semantic sizes. This mirrors human tendencies, where larger semantic sizes are often more engaging and attention-grabbing. However, this raises important considerations for the future. In a world where GUI agents and LLM-driven automation will become more prevalent, there is a possibility that clickbait headlines with exaggerated semantic sizes could be used to manipulate LLM agents, potentially compromising their safety and decision-making processes. The tendency for LLMs, particularly MLLMs, to emotionally engage with content of larger semantic size highlights the need for responsible design and ethical consideration in future AI systems to mitigate potential vulnerabilities. Ensuring these agents can make unbiased, well-informed decisions will be crucial as they take on more real-world responsibilities. While progress has been made, further refinement of multi-modal training and attention to the ethical dimensions of LLMs' emotional involvement are necessary.

\paragraph{Peering into Human Cognition: Insights on Multi-modal Learning from LLMs}
Our results suggest that multi-modal training enable LLMs to perceive and interact with the world, significantly enhancing their cognitive abilities. Due to the similarities between LLMs and humans, our findings from computational studies allow us to reasonably infer that human cognitive abilities are not acquired solely through language but require multi-modal interactions for experiential grounding.

In the real world, it is not feasible to conduct such modality-restrictive experiments on humans, where subjects are isolated to receive only language-based input, due to ethical constraints. However, through LLMs, we are able to simulate such conditions computationally, offering insights into how human cognitive development may rely on interactions with real-world through multiple modalities. This presents LLMs as valuable `brains in a vat' for exploring the role of real-world grounding experiences in shaping human cognition. As we continue to refine LLMs, these computational models offer a unique opportunity to study the foundational aspects of human cognitive development in ways that would otherwise be impossible.

\section{Conclusion}
Understanding human cognition has long been a complex challenge, with no single method able to fully capture its depth and intricacy. As large language models (LLMs) evolve, their increasing resemblance to human thought processes offers a promising new avenue for studying cognition, as they can, to some extent, simulate human cognitive abilities. Therefore, in this paper, we explored a key facet of cognition—semantic size—as a way to bridge the gap between human and artificial intelligence. We investigates how LLMs and humans perceive and understand semantic size, showing that multi-modal training significantly enhances their cognitive abilities. Through three studies—spanning metaphorical reasoning (Section \ref{sec:ex}), probing internal representations (Section \ref{sec:in}), and analyzing biases toward attention-grabbing content (Section \ref{sec:real})—we found that MLLMs are better equipped to grasp both abstract and concrete concepts. However, challenges remain, particularly in their understanding of abstract ideas, where even advanced models fall short of human-level cognition.

Our findings also suggest that multi-modal learning plays a pivotal role in enhancing LLMs' cognitive capacities, enabling them to engage more deeply with the world. These insights provide valuable clues into human cognition, reinforcing the idea that, like LLMs, human cognitive abilities are shaped not by language alone but through real-world engagement experiences. While it is not feasible to restrict human learning solely through language in the real-world, LLMs offer a unique window into understanding how multi-modal grounding experiences contribute to cognitive development. By simulating such conditions computationally, LLMs act as “brains in a vat,” helping us explore the essential role of real-world interaction in shaping cognition.
Moreover, our findings demonstrate that MLLMs are more drawn to titles with larger semantic sizes, reflecting human tendencies to engage with emotionally charged content. As LLMs, especially in their future roles as AI agents, become more involved in decision-making, it is essential to recognize how biases toward larger semantic sizes could influence their choices. Just as humans can be swayed by emotionally potent content, LLMs may also be vulnerable to manipulation through clickbait or other attention-grabbing tactics.

In essence, while LLMs are making strides in understanding semantic size, there is still much to explore.
As these models become more emotionally responsive, the ethical implications of their role in decision-making must be carefully considered. Ensuring that LLMs function safely and effectively in emotionally complex environments will be crucial as we continue to refine these models on the path toward human-like cognition. In the near future, LLMs may come to truly understand abstract emotions like love, narrowing the gap between human and artificial cognition in ways yet to be realized.

\section*{Inclusion and Ethics}
In this study, we ensured that our participants represented a diverse cross-section of society, encompassing various genders, age groups, and social backgrounds. All participants were recruited voluntarily through the online questionnaire platform, and each provided informed consent before participating. They were thoroughly briefed about the purpose of the research, ensuring transparency and understanding.
Furthermore, the domains used in our experiments and the examples included in this paper do not pose any specific ethical concerns or risks to participants. Our study involved providing examples and prompts in both English and the participants' native languages, ensuring ease of understanding and inclusivity throughout the process. 



\newpage

\bibliography{sn-bibliography}

\appendix
\section{The Glasgow Norms}
\label{append:glasgow}

The detailed descriptions and labels of the Concreteness and Semantic Size are shown in Table \ref{tab:Glasgow}
\begin{table}[h]
\centering
\begin{tabular}{@{}ll@{}}
\toprule
\textbf{Concreteness}  &                                                                                                                                                                              \\ \midrule
Description            & Concreteness is a measure of how concrete or abstract something is. A word is \\
                       & CONCRETE if it represents something that exists in a definite physical form \\
                       & in the real world. In contrast, a word is ABSTRACT if it represents more of a \\
                       & concept or idea.\\
Labels                 & 1-7                                                                                                                                                                          \\ \midrule
\textbf{Semantic Size} &                                                                                                                                                                              \\ \midrule
Description            & Size is a measure of something’s dimensions, magnitude, or extent. A word \\
                       & represents something BIG if it refers to things or concepts that are large. A\\
                       & word represents something SMALL if it refers to things or concepts that are little.                                                                                        \\
Labels                 & 1-7                                                                                                                                                                          \\ \bottomrule
\end{tabular}
\caption{The Glasgow Norms's description and labels of Concreteness and Semantic Size}
\label{tab:Glasgow}
\end{table}

\section{Semantic Size Metaphor Dataset}
\label{append:metaphor_dataset}
\begin{figure}
    \centering
    \includegraphics[width=1\linewidth]{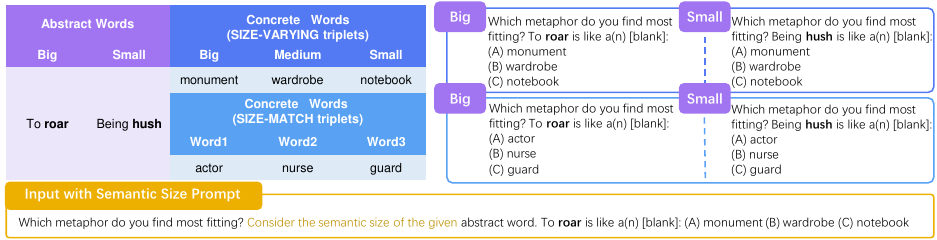}
    \caption{Detailed examples for Semantic Size Metaphor dataset with multiple-choice question format and semantic size prompt format}
    \label{fig:dataset_large}
\end{figure}

The detailed examples for Semantic Size Metaphor dataset with multiple-choice question format are illustrated in the Figure \ref{fig:dataset_large}.

\section{Participant Information}
\label{append:participant}
\begin{figure}
    \centering
    \includegraphics[width=1\linewidth]{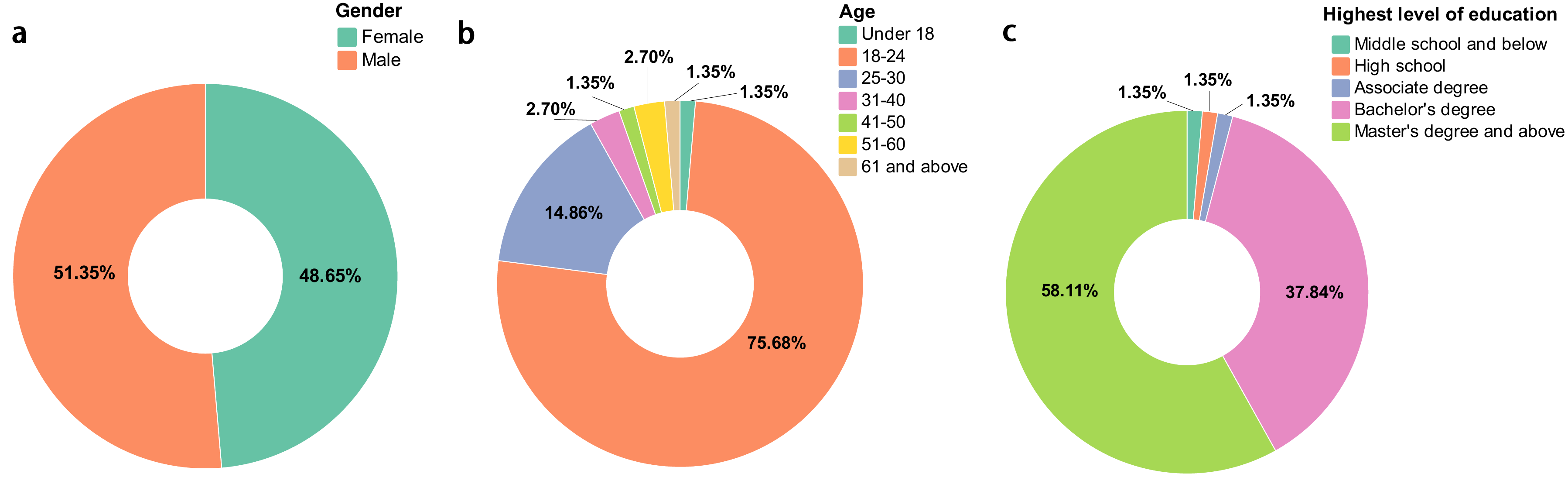}
    \caption{Detailed statistics of participants, including gender (a), age (b), and the highest level of eduacation (c)}
    \label{fig:human_eval_sta}
\end{figure}

For the human evaluation, we recruited 74 participants. All participants had normal or corrected-to-normal vision and had no history of learning or language disorders (e.g., dyslexia). The experiment lasted approximately 20 minutes. Detailed pie chart information about the participants, including gender, age, and educational background, is provided in Figure \ref{fig:human_eval_sta}.

\section{Further Exploration}
\label{append:semantic size prompt}
\subsection{Results with semantic size prompt}
To further investigate whether LLMs can make metaphorical associations based on semantic size, we simplified the experiment by directly prompting the models with, `Consider the semantic size of the given abstract word.' The detailed input format can be found in Figure \ref{fig:dataset_large}  (yellow box), and the overall results are presented in Table \ref{tab:exp1_extend_prompt_size}. As seen in the table, most models demonstrate a tendency more closely aligned with human reasoning after receiving this explicit prompt. However, some models, such as Yi-VL-6B and Qwen-VL-Chat, still struggle to fully grasp semantic size, even with direct prompting.

We also conducted experiments using the original, unexpanded dataset constructed by \cite{yao2022can}. Detailed results of these experiments are provided in Table \ref{tab:exp1_unexpanded} and Tabel \ref{tab:exp1_unexpanded_prompt_size}, which show the outcomes without the semantic size prompt and with the semantic size prompt, respectively.

\begin{table}[]
\centering
\scalebox{0.9}{
\begin{tabular}{@{}llrrrrrrrr@{}}
\toprule
\multirow{2}{*}{Model}                    & \multirow{2}{*}{Size} & \multicolumn{4}{c}{Size-Varying Concrete Triplets}                                                                                          & \multicolumn{4}{c}{Size-Match Concrete Triplets}                                                                             \\ \cmidrule(l){3-10} 
                                          &                       & \multicolumn{1}{c}{big} & \multicolumn{1}{c}{middle} & \multicolumn{1}{c}{small} & \multicolumn{1}{c|}{\textbf{C}} & \multicolumn{1}{c}{0} & \multicolumn{1}{c}{1} & \multicolumn{1}{c}{2} & \multicolumn{1}{c}{\textbf{C}} \\ \midrule
GPT-3.5-turbo \cite{gpt4}                             & Big                   & \textbf{63.28}          & 18.59                      & 18.13                     & \multicolumn{1}{r|}{29.95}     & \textbf{37.50}        & 35.00                 & 27.50                 & 4.17                          \\
                                          & Small                 & 25.16                   & 17.97                      & \textbf{56.88}            & \multicolumn{1}{r|}{23.54}     & \textbf{35.31}        & 32.81                 & 31.88                 & 1.98                          \\
\multirow{2}{*}{GPT-4-turbo \cite{gpt4}}              & Big                   & \textbf{69.84}          & 17.66                      & 12.50                     & \multicolumn{1}{r|}{36.51}     & \textbf{37.03}        & 30.47                 & 32.50                 & 3.70                          \\
                                          & Small                 & 32.81                   & 25.63                      & \textbf{41.56}            & \multicolumn{1}{r|}{8.23}      & 33.13                 & \textbf{34.22}        & 32.66                 & 0.89                          \\
\multirow{2}{*}{GPT-4o \cite{gpt4}}                   & Big                   & \textbf{70.47}          & 15.94                      & 13.59                     & \multicolumn{1}{r|}{37.14}     & \textbf{37.81}        & 32.97                 & 29.22                 & 4.48                          \\
                                          & Small                 & 16.56                   & 17.03                      & \textbf{66.41}            & \multicolumn{1}{r|}{33.07}     & 34.53                 & \textbf{36.25}        & 29.22                 & 2.92                          \\
\multirow{2}{*}{GPT-4o-mini \cite{gpt4}}              & Big                   & \textbf{69.22}          & 17.66                      & 13.13                     & \multicolumn{1}{r|}{35.89}     & \textbf{36.88}        & 32.81                 & 30.31                 & 3.54                          \\
                                          & Small                 & 29.38                   & 23.91                      & \textbf{46.72}            & \multicolumn{1}{r|}{13.39}     & \textbf{37.81}        & 30.47                 & 31.72                 & 4.48                          \\ \midrule
\multirow{2}{*}{Yi-6B-Chat \cite{ai2024yi}}               & Big                   & \textbf{43.44}          & 26.25                      & 30.31                     & \multicolumn{1}{r|}{10.10}     & \textbf{36.72}        & 33.28                 & 30.00                 & 3.39                          \\
                                          & Small                 & 34.30                   & 27.42                      & \textbf{38.28}            & \multicolumn{1}{r|}{4.95}      & \textbf{35.08}        & 32.50                 & \textbf{32.42}        & 1.74                          \\
\multirow{2}{*}{Yi-VL-6B \cite{ai2024yi}}                 & Big                   & \textbf{54.92}          & 21.17                      & 23.91                     & \multicolumn{1}{r|}{21.59}     & \textbf{35.55}        & 33.13                 & 31.33                 & 2.21                          \\
                                          & Small                 & \textbf{39.22}          & 23.28                      & 37.50                     & \multicolumn{1}{r|}{5.89}      & 31.95                 & \textbf{38.67}        & 29.38                 & 5.34                          \\ \midrule
\multirow{2}{*}{Yi-34B-Chat \cite{ai2024yi}}              & Big                   & \textbf{67.11}          & 19.30                      & 13.59                     & \multicolumn{1}{r|}{33.78}     & \textbf{36.02}        & 34.14                 & 29.84                 & 2.68                          \\
                                          & Small                 & \textbf{60.63}          & 19.53                      & 19.84                     & \multicolumn{1}{r|}{27.29}     & \textbf{37.81}        & 34.22                 & 27.97                 & 4.48                          \\
\multirow{2}{*}{Yi-VL-34B \cite{ai2024yi}}                & Big                   & \textbf{55.94}          & 22.19                      & 21.88                     & \multicolumn{1}{r|}{22.60}     & 32.42                 & 33.36                 & \textbf{34.22}        & 0.89                          \\
                                          & Small                 & 26.72                   & 21.72                      & \textbf{51.56}            & \multicolumn{1}{r|}{18.23}     & \textbf{35.55}        & 34.14                 & 30.31                 & 2.21                          \\ \midrule
                                          & Small                 & 31.56                   & 23.91                      & \textbf{44.53}            & \multicolumn{1}{r|}{11.20}     & 34.92                 & \textbf{39.45}        & 25.63                 & 6.12                          \\
\multirow{2}{*}{Qwen-7B-Chat \cite{qwen}}             & Big                   & \textbf{43.28}          & 33.52                      & 23.20                     & \multicolumn{1}{r|}{9.95}      & \textbf{36.25}        & 31.56                 & 32.19                 & 2.92                          \\
                                          & Small                 & \textbf{45.31}          & 30.94                      & 23.75                     & \multicolumn{1}{r|}{11.98}     & \textbf{34.61}        & 31.80                 & 33.59                 & 1.28                          \\
\multirow{2}{*}{Qwen-VL-Chat \cite{qwen}}             & Big                   & \textbf{64.38}          & 19.69                      & 15.94                     & \multicolumn{1}{r|}{31.04}     & \textbf{36.88}        & 33.36                 & 29.77                 & 3.54                          \\
                                          & Small                 & \textbf{54.69}          & 25.39                      & 19.92                     & \multicolumn{1}{r|}{21.35}     & \textbf{38.13}        & 31.02                 & 30.86                 & 4.79                          \\
\multirow{2}{*}{Qwen-VL-max \cite{qwen}}              & Big                   & \textbf{48.28}          & 20.94                      & 30.78                     & \multicolumn{1}{r|}{14.95}     & \textbf{39.61}        & 28.52                 & 31.88                 & 6.28                          \\
                                          & Small                 & 21.88                   & 24.61                      & \textbf{53.52}            & \multicolumn{1}{r|}{20.18}     & \textbf{38.98}        & 29.77                 & 31.25                 & 5.65                          \\
\multirow{2}{*}{Qwen-VL-plus \cite{qwen}}             & Big                   & \textbf{46.25}          & 25.23                      & 28.52                     & \multicolumn{1}{r|}{12.92}     & \textbf{37.97}        & 33.67                 & 28.36                 & 4.64                          \\
                                          & Small                 & 27.97                   & 27.73                      & \textbf{44.30}            & \multicolumn{1}{r|}{10.96}     & 34.30                 & \textbf{36.02}        & 29.69                 & 2.68                          \\ \midrule
\multirow{2}{*}{Mistral-7B-Instruct-v0.2 \cite{jiang2023mistral}} & Big                   & \textbf{47.50}          & 30.08                      & 22.42                     & \multicolumn{1}{r|}{14.17}     & 34.61                 & \textbf{36.72}        & 28.67                 & 3.39                          \\
                                          & Small                 & 31.09                   & 30.31                      & \textbf{38.59}            & \multicolumn{1}{r|}{5.26}      & \textbf{36.64}        & 33.59                 & 29.77                 & 3.31                          \\
\multirow{2}{*}{Llava-v1.6-Mistral-7b-hf \cite{llava}} & Big                   & \textbf{50.55}          & 29.22                      & 20.23                     & \multicolumn{1}{r|}{17.21}     & \textbf{35.23}        & 33.91                 & 30.86                 & 1.90                          \\
                                          & Small                 & \textbf{40.70}          & 29.06                      & 30.23                     & \multicolumn{1}{r|}{7.37}      & \textbf{38.52}        & 36.17                 & 25.31                 & 5.18                          \\ \midrule
\multirow{2}{*}{Vicuna-7b-v1.5 \cite{vicuna}}           & Big                   & \textbf{34.61}          & \textbf{34.61}             & 30.78                     & \multicolumn{1}{r|}{1.28}      & 31.33                 & 33.91                 & \textbf{34.77}        & 1.43                          \\
                                          & Small                 & \textbf{34.45}          & 34.14                      & 31.41                     & \multicolumn{1}{r|}{1.12}      & 30.86                 & 33.67                 & \textbf{35.47}        & 2.14                          \\
\multirow{2}{*}{Llava-v1.6-Vicuna-7b-hf \cite{llava}}  & Big                   & \textbf{42.19}          & 28.44                      & 29.38                     & \multicolumn{1}{r|}{8.85}      & 33.75                 & \textbf{34.22}        & 32.03                 & 0.89                          \\
                                          & Small                 & 32.58                   & 32.11                      & \textbf{35.31}            & \multicolumn{1}{r|}{1.98}      & \textbf{36.25}        & 31.88                 & 31.88                 & 2.92                          \\ \midrule
\multirow{2}{*}{Vicuna-13b-v1.5 \cite{vicuna}}          & Big                   & \textbf{35.78}          & 33.91                      & 30.31                     & \multicolumn{1}{r|}{2.45}      & \textbf{36.41}        & 33.67                 & 29.92                 & 3.07                          \\
                                          & Small                 & 32.19                   & \textbf{35.23}             & 32.58                     & \multicolumn{1}{r|}{1.90}      & \textbf{35.86}        & 35.23                 & 28.91                 & 2.53                          \\
\multirow{2}{*}{Llava-v1.6-Vicuna-13b-hf \cite{llava}} & Big                   & \textbf{44.77}          & 25.23                      & 30.00                     & \multicolumn{1}{r|}{11.43}     & 31.72                 & 33.59                 & \textbf{34.69}        & 1.35                          \\
                                          & Small                 & 29.45                   & 27.89                      & \textbf{42.66}            & \multicolumn{1}{r|}{9.32}      & \textbf{34.61}        & 34.06                 & 31.33                 & 1.28                          \\ \midrule
\multirow{2}{*}{Meta-Llama-3-8B-Instruct \cite{llama3modelcard}} & Big                   & \textbf{34.30}          & 31.41                      & \textbf{34.30}            & \multicolumn{1}{r|}{0.96}      & 31.33                 & 34.22                 & \textbf{34.45}        & 1.12                          \\
                                          & Small                 & 33.13                   & 31.80                      & \textbf{35.08}            & \multicolumn{1}{r|}{1.74}      & 31.33                 & 34.06                 & \textbf{34.61}        & 1.28                          \\
\multirow{2}{*}{CogVLM2-llama3-Chat-19B \cite{hong2024cogvlm2}}  & Big                   & \textbf{35.31}          & 30.94                      & 33.75                     & \multicolumn{1}{r|}{1.98}      & \textbf{38.20}        & 32.11                 & 29.69                 & 4.87                          \\
                                          & Small                 & 31.17                   & 29.61                      & \textbf{39.22}            & \multicolumn{1}{r|}{5.89}      & \textbf{36.72}        & 33.05                 & 30.23                 & 3.39                          \\
\multirow{2}{*}{MiniCPM-Llama3-V-2\_5 \cite{yao2024minicpm}}    & Big                   & \textbf{56.41}          & 20.16                      & 23.44                     & \multicolumn{1}{r|}{23.07}     & \textbf{41.95}        & 30.39                 & 27.66                 & 8.62                          \\
                                          & Small                 & 24.45                   & 21.64                      & \textbf{53.91}            & \multicolumn{1}{r|}{20.57}     & \textbf{38.36}        & 32.81                 & 28.83                 & 5.03                          \\ \bottomrule
\end{tabular}}
\caption{Results for Experiment 1 on the extended dataset with semantic size prompt}
\label{tab:exp1_extend_prompt_size}
\end{table}

\begin{table}[]
\centering
\scalebox{0.9}{
\begin{tabular}{@{}lllllrlllr@{}}
\toprule
\multirow{2}{*}{Model}                    & \multirow{2}{*}{Size} & \multicolumn{4}{c}{Size-Varying Concrete Triplets}                                                                                                              & \multicolumn{4}{c}{Size-Match Concrete Triplets}                                                                                                                    \\ \cmidrule(l){3-10} 
                                          &                       & \multicolumn{1}{c}{big}            & \multicolumn{1}{c}{middle} & \multicolumn{1}{c}{small}          & \multicolumn{1}{c|}{\textbf{C}} & \multicolumn{1}{c}{0}              & \multicolumn{1}{c}{1}              & \multicolumn{1}{c}{2}              & \multicolumn{1}{c}{\textbf{C}} \\ \midrule
\multirow{2}{*}{Human}                    & Big                   & \multicolumn{1}{c}{\textbf{49.00}} & \multicolumn{1}{c}{26.00}  & \multicolumn{1}{c}{25.00}          & \multicolumn{1}{r|}{15.67}     & \multicolumn{1}{c}{\textbf{34.00}} & \multicolumn{1}{c}{\textbf{34.00}} & \multicolumn{1}{c}{32.00}          & 5.67                          \\
                                          & Small                 & \multicolumn{1}{c}{30.00}          & \multicolumn{1}{c}{25.00}  & \multicolumn{1}{c}{\textbf{46.00}} & \multicolumn{1}{r|}{12.33}     & \multicolumn{1}{c}{31.00}          & \multicolumn{1}{c}{33.00}          & \multicolumn{1}{c}{\textbf{36.00}} & 12.00                         \\ \midrule
GPT-3.5-turbo \cite{gpt4}                             & Big                   & \textbf{50.18}                     & 20.73                      & 29.09                              & \multicolumn{1}{r|}{16.85}     & \textbf{37.82}                     & 36.73                              & 25.45                              & 4.48                          \\
                                          & Small                 & 28.00                              & 29.09                      & \textbf{42.91}                     & \multicolumn{1}{r|}{9.58}      & \textbf{38.91}                     & 31.64                              & 29.45                              & 5.58                          \\
\multirow{2}{*}{GPT-4-turbo \cite{gpt4}}              & Big                   & \textbf{57.09}                     & 19.64                      & 23.27                              & \multicolumn{1}{r|}{23.76}     & \textbf{39.27}                     & 34.18                              & 26.55                              & 5.94                          \\
                                          & Small                 & \textbf{43.64}                     & 26.91                      & 29.45                              & \multicolumn{1}{r|}{10.30}     & 34.18                              & \textbf{34.91}                     & 30.91                              & 1.58                          \\
\multirow{2}{*}{GPT-4o \cite{gpt4}}                   & Big                   & \textbf{49.82}                     & 18.91                      & 31.27                              & \multicolumn{1}{r|}{16.48}     & \textbf{39.64}                     & 37.45                              & 22.91                              & 6.30                          \\
                                          & Small                 & 31.27                              & 26.18                      & \textbf{42.55}                     & \multicolumn{1}{r|}{9.21}      & 33.09                              & \textbf{34.91}                     & 32.00                              & 1.58                          \\
\multirow{2}{*}{GPT-4o-mini \cite{gpt4}}              & Big                   & \textbf{60.36}                     & 14.91                      & 24.73                              & \multicolumn{1}{r|}{27.03}     & 34.18                              & \textbf{42.91}                     & 22.91                              & 9.58                          \\
                                          & Small                 & 26.55                              & 28.00                      & \textbf{45.45}                     & \multicolumn{1}{r|}{12.12}     & 33.09                              & \textbf{33.82}                     & 33.09                              & 0.48                          \\ \midrule
\multirow{2}{*}{Yi-6B-Chat \cite{ai2024yi}}               & Big                   & \textbf{41.45}                     & 25.64                      & 32.91                              & \multicolumn{1}{r|}{8.12}      & 28.00                              & \textbf{40.91}                     & 31.09                              & 7.58                          \\
                                          & Small                 & 31.27                              & 28.00                      & \textbf{40.73}                     & \multicolumn{1}{r|}{7.39}      & 30.73                              & 32.91                              & \textbf{36.36}                     & 3.03                          \\
\multirow{2}{*}{Yi-VL-6B \cite{ai2024yi}}                 & Big                   & \textbf{47.45}                     & 18.91                      & 33.64                              & \multicolumn{1}{r|}{14.12}     & 30.00                              & \textbf{40.55}                     & 29.45                              & 7.21                          \\
                                          & Small                 & 29.82                              & 30.18                      & \textbf{40.00}                     & \multicolumn{1}{r|}{6.67}      & \textbf{34.36}                     & 31.27                              & \textbf{34.36}                     & 1.03                          \\ \midrule
\multirow{2}{*}{Yi-34B-Chat \cite{ai2024yi}}              & Big                   & \textbf{50.55}                     & 24.00                      & 25.45                              & \multicolumn{1}{r|}{17.21}     & 34.73                              & 38.55                              & 26.73                              & 5.21                          \\
                                          & Small                 & 36.73                              & 24.55                      & \textbf{38.73}                     & \multicolumn{1}{r|}{5.39}      & 36.73                              & 33.82                              & 29.45                              & 3.39                          \\
\multirow{2}{*}{Yi-VL-34B \cite{ai2024yi}}                & Big                   & \textbf{42.18}                     & 30.36                      & 27.45                              & \multicolumn{1}{r|}{8.85}      & 30.55                              & 43.82                              & 25.64                              & 10.48                         \\
                                          & Small                 & 23.82                              & 33.64                      & \textbf{42.55}                     & \multicolumn{1}{r|}{9.21}      & 34.91                              & 34.55                              & 30.55                              & 1.58                          \\ \midrule
\multirow{2}{*}{Qwen1.5-7B-Chat \cite{qwen}}          & Big                   & \textbf{55.45}                     & 18.91                      & 25.64                              & \multicolumn{1}{r|}{22.12}     & 30.00                              & \textbf{44.36}                     & 25.64                              & 11.03                         \\
                                          & Small                 & 28.91                              & 30.36                      & \textbf{40.73}                     & \multicolumn{1}{r|}{7.39}      & 31.64                              & \textbf{40.18}                     & 28.18                              & 6.85                          \\
\multirow{2}{*}{Qwen-7B-Chat \cite{qwen}}             & Big                   & \textbf{45.09}                     & 27.27                      & 27.64                              & \multicolumn{1}{r|}{11.76}     & \textbf{35.09}                     & 30.55                              & 34.36                              & 1.76                          \\
                                          & Small                 & \textbf{43.27}                     & 30.73                      & 26.00                              & \multicolumn{1}{r|}{9.94}      & 34.36                              & 30.00                              & \textbf{35.64}                     & 2.30                          \\
\multirow{2}{*}{Qwen-VL-Chat \cite{qwen}}             & Big                   & \textbf{51.82}                     & 22.36                      & 25.82                              & \multicolumn{1}{r|}{18.48}     & 32.91                              & \textbf{37.64}                     & 29.45                              & 4.30                          \\
                                          & Small                 & \textbf{48.18}                     & 25.27                      & 26.55                              & \multicolumn{1}{r|}{14.85}     & \textbf{35.82}                     & 31.09                              & 33.09                              & 2.48                          \\
\multirow{2}{*}{Qwen-VL-max \cite{qwen}}              & Big                   & 36.18                              & 26.18                      & \textbf{37.64}                     & \multicolumn{1}{r|}{4.30}      & \textbf{37.64}                     & 39.45                              & 22.91                              & 6.12                          \\
                                          & Small                 & 20.00                              & 28.36                      & \textbf{51.64}                     & \multicolumn{1}{r|}{18.30}     & 35.09                              & 29.27                              & \textbf{35.64}                     & 2.30                          \\
\multirow{2}{*}{Qwen-VL-plus \cite{qwen}}             & Big                   & \textbf{43.45}                     & 23.82                      & 32.73                              & \multicolumn{1}{r|}{10.12}     & 34.36                              & \textbf{39.45}                     & 26.18                              & 6.12                          \\
                                          & Small                 & 28.36                              & 31.82                      & \textbf{39.82}                     & \multicolumn{1}{r|}{6.48}      & \textbf{35.64}                     & 32.18                              & 32.18                              & 2.30                          \\ \midrule
\multirow{2}{*}{Mistral-7B-Instruct-v0.2 \cite{jiang2023mistral}} & Big                   & \textbf{43.45}                     & 23.27                      & 33.27                              & \multicolumn{1}{r|}{10.12}     & 30.73                              & \textbf{39.09}                     & 30.18                              & 5.76                          \\
                                          & Small                 & 26.00                              & 27.09                      & \textbf{46.91}                     & \multicolumn{1}{r|}{13.58}     & \textbf{39.09}                     & 33.09                              & 27.82                              & 5.76                          \\
\multirow{2}{*}{Llava-v1.6-Mistral-7b-hf \cite{llava}} & Big                   & \textbf{43.64}                     & 24.18                      & 32.18                              & \multicolumn{1}{r|}{10.30}     & 30.36                              & \textbf{38.18}                     & 31.45                              & 4.85                          \\
                                          & Small                 & 24.18                              & 31.09                      & \textbf{44.73}                     & \multicolumn{1}{r|}{11.39}     & \textbf{38.55}                     & 31.82                              & 29.64                              & 5.21                          \\ \midrule
\multirow{2}{*}{Vicuna-7b-v1.5 \cite{vicuna}}           & Big                   & \textbf{34.36}                     & 33.45                      & 32.18                              & \multicolumn{1}{r|}{1.03}      & 30.91                              & 32.36                              & \textbf{36.73}                     & 3.39                          \\
                                          & Small                 & 33.09                              & \textbf{34.36}             & 32.55                              & \multicolumn{1}{r|}{1.03}      & 32.18                              & \textbf{34.00}                     & 33.82                              & 0.67                          \\
\multirow{2}{*}{Llava-v1.6-Vicuna-7b-hf \cite{llava}}  & Big                   & 35.27                              & 28.73                      & \textbf{36.00}                     & \multicolumn{1}{r|}{2.67}      & 31.64                              & \textbf{39.27}                     & 29.09                              & 5.94                          \\
                                          & Small                 & 28.91                              & 32.18                      & \textbf{38.91}                     & \multicolumn{1}{r|}{5.58}      & 33.27                              & 32.91                              & \textbf{33.82}                     & 0.48                          \\ \midrule
\multirow{2}{*}{Vicuna-13b-v1.5 \cite{vicuna}}          & Big                   & \textbf{36.00}                     & 32.73                      & 31.27                              & \multicolumn{1}{r|}{2.67}      & \textbf{36.36}                     & 33.27                              & 30.36                              & 3.03                          \\
                                          & Small                 & \textbf{35.45}                     & 35.27                      & 29.27                              & \multicolumn{1}{r|}{2.12}      & \textbf{35.82}                     & 34.00                              & 30.18                              & 2.48                          \\
\multirow{2}{*}{Llava-v1.6-Vicuna-13b-hf \cite{llava}} & Big                   & \textbf{42.55}                     & 22.00                      & 35.45                              & \multicolumn{1}{r|}{9.21}      & 29.09                              & \textbf{36.36}                     & 34.55                              & 3.03                          \\
                                          & Small                 & 26.55                              & 31.45                      & \textbf{42.00}                     & \multicolumn{1}{r|}{8.67}      & \textbf{35.64}                     & 31.09                              & 33.27                              & 2.30                          \\ \midrule
\multirow{2}{*}{Meta-Llama-3-8B-Instruct \cite{llama3modelcard}} & Big                   & \textbf{36.36}                     & 30.36                      & 33.27                              & \multicolumn{1}{r|}{3.03}      & \textbf{34.18}                     & 33.64                              & 32.18                              & 0.85                          \\
                                          & Small                 & \textbf{34.55}                     & 31.45                      & 34.00                              & \multicolumn{1}{r|}{1.21}      & \textbf{36.18}                     & 31.27                              & 32.55                              & 2.85                          \\
\multirow{2}{*}{CogVLM2-llama3-Chat-19B \cite{hong2024cogvlm2}}  & Big                   & \textbf{40.00}                     & 27.09                      & 32.91                              & \multicolumn{1}{r|}{6.67}      & \textbf{38.18}                     & 31.09                              & 30.73                              & 4.85                          \\
                                          & Small                 & 28.36                              & 32.18                      & \textbf{39.45}                     & \multicolumn{1}{r|}{6.12}      & \textbf{35.82}                     & 32.55                              & 31.64                              & 2.48                          \\
\multirow{2}{*}{MiniCPM-Llama3-V-2\_5 \cite{yao2024minicpm}}    & Big                   & \textbf{52.36}                     & 20.73                      & 26.91                              & \multicolumn{1}{r|}{19.03}     & \textbf{43.27}                     & 34.18                              & 22.55                              & 9.94                          \\
                                          & Small                 & 28.00                              & 28.36                      & \textbf{43.64}                     & \multicolumn{1}{r|}{10.30}     & 30.55                              & \textbf{36.00}                     & 33.45                              & 2.67                          \\ \bottomrule
\end{tabular}
}
\caption{Results for Experiment 1 on the unexpanded dataset}
\label{tab:exp1_unexpanded}
\end{table}

\begin{table}[]
\centering
\scalebox{0.9}{
\begin{tabular}{@{}llrrrrrrrr@{}}
\toprule
\multirow{2}{*}{Model}                    & \multirow{2}{*}{Size} & \multicolumn{4}{c}{Size-Varying Concrete Triplets}                                                                                          & \multicolumn{4}{c}{Size-Match Concrete Triplets}                                                                             \\ \cmidrule(l){3-10} 
                                          &                       & \multicolumn{1}{c}{big} & \multicolumn{1}{c}{middle} & \multicolumn{1}{c}{small} & \multicolumn{1}{c|}{\textbf{C}} & \multicolumn{1}{c}{0} & \multicolumn{1}{c}{1} & \multicolumn{1}{c}{2} & \multicolumn{1}{c}{\textbf{C}} \\ \midrule
GPT-3.5-turbo \cite{gpt4}                             & Big                   & \textbf{67.64}          & 16.00                      & 16.36                     & \multicolumn{1}{r|}{34.30}     & \textbf{38.18}        & 36.36                 & 25.45                 & 4.85                          \\
                                          & Small                 & 32.73                   & 22.55                      & \textbf{44.73}            & \multicolumn{1}{r|}{11.39}     & 33.45                 & 30.18                 & \textbf{36.36}        & 3.03                          \\
\multirow{2}{*}{GPT-4-turbo \cite{gpt4} }              & Big                   & \textbf{68.36}          & 14.91                      & 16.73                     & \multicolumn{1}{r|}{35.03}     & \textbf{42.18}        & 28.73                 & 29.09                 & 8.85                          \\
                                          & Small                 & 36.73                   & 23.27                      & \textbf{40.00}            & \multicolumn{1}{r|}{6.67}      & \textbf{36.73}        & 32.00                 & 31.27                 & 3.39                          \\
\multirow{2}{*}{GPT-4o \cite{gpt4} }                   & Big                   & \textbf{75.64}          & 9.09                       & 15.27                     & \multicolumn{1}{r|}{42.30}     & \textbf{39.64}        & 29.45                 & 30.91                 & 6.30                          \\
                                          & Small                 & 10.55                   & 15.27                      & \textbf{74.18}            & \multicolumn{1}{r|}{40.85}     & 30.18                 & \textbf{35.27}        & 34.55                 & 1.94                          \\
\multirow{2}{*}{GPT-4o-mini \cite{gpt4} }              & Big                   & \textbf{73.82}          & 10.91                      & 15.27                     & \multicolumn{1}{r|}{40.48}     & \textbf{37.82}        & \textbf{37.82}        & 24.36                 & 4.48                          \\
                                          & Small                 & 25.09                   & 29.82                      & \textbf{45.09}            & \multicolumn{1}{r|}{11.76}     & \textbf{38.91}        & 30.91                 & 30.18                 & 5.58                          \\ \midrule
\multirow{2}{*}{Yi-6B-Chat \cite{ai2024yi}}               & Big                   & \textbf{42.55}          & 23.27                      & 34.18                     & \multicolumn{1}{r|}{9.21}      & 29.27                 & \textbf{37.27}        & 33.45                 & 3.94                          \\
                                          & Small                 & 30.55                   & 24.00                      & \textbf{45.45}            & \multicolumn{1}{r|}{12.12}     & \textbf{35.45}        & 29.09                 & \textbf{35.45}        & 2.12                          \\
\multirow{2}{*}{Yi-VL-6B \cite{ai2024yi}}                 & Big                   & \textbf{57.09}          & 14.91                      & 28.00                     & \multicolumn{1}{r|}{23.76}     & 32.36                 & \textbf{41.09}        & 26.55                 & 7.76                          \\
                                          & Small                 & 32.36                   & 25.64                      & \textbf{42.00}            & \multicolumn{1}{r|}{8.67}      & \textbf{35.09}        & 30.91                 & 34.00                 & 1.76                          \\ \midrule
\multirow{2}{*}{Yi-34B-Chat \cite{ai2024yi}}              & Big                   & \textbf{69.64}          & 13.45                      & 16.91                     & \multicolumn{1}{r|}{36.30}     & 34.36                 & \textbf{38.36}        & 27.27                 & 5.03                          \\
                                          & Small                 & \textbf{55.82}          & 20.18                      & 24.00                     & \multicolumn{1}{r|}{22.48}     & \textbf{39.27}        & 30.91                 & 29.82                 & 5.94                          \\
\multirow{2}{*}{Yi-VL-34B \cite{ai2024yi}}                & Big                   & \textbf{56.36}          & 22.18                      & 21.45                     & \multicolumn{1}{r|}{23.03}     & 33.45                 & \textbf{38.18}        & 28.36                 & 4.85                          \\
                                          & Small                 & 22.91                   & 24.18                      & \textbf{52.91}            & \multicolumn{1}{r|}{19.58}     & \textbf{36.18}        & 35.27                 & 28.55                 & 2.85                          \\ \midrule
\multirow{2}{*}{Qwen1.5-7B-Chat \cite{qwen}}          & Big                   & \textbf{59.45}          & 17.82                      & 22.73                     & \multicolumn{1}{r|}{26.12}     & 32.91                 & \textbf{38.36}        & 28.73                 & 5.03                          \\
                                          & Small                 & 28.18                   & 31.45                      & \textbf{40.36}            & \multicolumn{1}{r|}{7.03}      & 34.73                 & \textbf{39.82}        & 25.45                 & 6.48                          \\
\multirow{2}{*}{Qwen-7B-Chat \cite{qwen}}             & Big                   & \textbf{45.64}          & 28.36                      & 26.00                     & \multicolumn{1}{r|}{12.30}     & \textbf{38.91}        & 34.00                 & 27.09                 & 5.58                          \\
                                          & Small                 & \textbf{43.64}          & 27.64                      & 28.73                     & \multicolumn{1}{r|}{10.30}     & 41.82                 & 30.36                 & \textbf{27.82}        & 8.48                          \\
\multirow{2}{*}{Qwen-VL-Chat \cite{qwen}}             & Big                   & \textbf{64.00}          & 17.27                      & 18.73                     & \multicolumn{1}{r|}{30.67}     & \textbf{38.00}        & 35.45                 & 26.55                 & 4.67                          \\
                                          & Small                 & \textbf{54.55}          & 26.73                      & 18.73                     & \multicolumn{1}{r|}{21.21}     & \textbf{37.09}        & 29.82                 & 33.09                 & 3.76                          \\
\multirow{2}{*}{Qwen-VL-max \cite{qwen}}              & Big                   & \textbf{51.27}          & 18.55                      & 30.18                     & \multicolumn{1}{r|}{17.94}     & \textbf{40.73}        & 38.18                 & 21.09                 & 7.39                          \\
                                          & Small                 & 22.91                   & 17.82                      & \textbf{59.27}            & \multicolumn{1}{r|}{25.94}     & \textbf{41.64}        & 31.09                 & 27.27                 & 8.30                          \\
\multirow{2}{*}{Qwen-VL-plus \cite{qwen}}             & Big                   & \textbf{52.36}          & 16.73                      & 30.91                     & \multicolumn{1}{r|}{19.03}     & 34.91                 & \textbf{41.64}        & 23.45                 & 8.30                          \\
                                          & Small                 & 28.18                   & 29.45                      & \textbf{42.36}            & \multicolumn{1}{r|}{9.03}      & 30.55                 & \textbf{40.18}        & 29.27                 & 6.85                          \\ \midrule
\multirow{2}{*}{Mistral-7B-Instruct-v0.2 \cite{jiang2023mistral}} & Big                   & \textbf{47.64}          & 24.55                      & 27.82                     & \multicolumn{1}{r|}{14.30}     & 36.55                 & \textbf{38.91}        & 24.55                 & 5.58                          \\
                                          & Small                 & 19.45                   & 32.91                      & \textbf{47.64}            & \multicolumn{1}{r|}{14.30}     & \textbf{39.64}        & 32.00                 & 28.36                 & 6.30                          \\
\multirow{2}{*}{Llava-v1.6-Mistral-7b-hf \cite{llava}} & Big                   & \textbf{52.00}          & 24.36                      & 23.64                     & \multicolumn{1}{r|}{18.67}     & \textbf{38.18}        & 37.64                 & 24.18                 & 4.85                          \\
                                          & Small                 & 35.09                   & 28.73                      & \textbf{36.18}            & \multicolumn{1}{r|}{2.85}      & \textbf{38.36}        & 38.00                 & 23.64                 & 5.03                          \\ \midrule
\multirow{2}{*}{Vicuna-7b-v1.5 \cite{vicuna}}           & Big                   & 35.45                   & 28.18                      & \textbf{36.36}            & \multicolumn{1}{r|}{3.03}      & 32.00                 & 33.09                 & \textbf{34.91}        & 1.58                          \\
                                          & Small                 & 34.18                   & 29.82                      & \textbf{36.00}            & \multicolumn{1}{r|}{2.67}      & 31.27                 & 31.64                 & \textbf{37.09}        & 3.76                          \\
\multirow{2}{*}{Llava-v1.6-Vicuna-7b-hf \cite{llava}}  & Big                   & \textbf{45.09}          & 29.82                      & 25.09                     & \multicolumn{1}{r|}{11.76}     & \textbf{33.64}        & 32.91                 & 33.45                 & 0.30                          \\
                                          & Small                 & 31.27                   & 31.64                      & \textbf{37.09}            & \multicolumn{1}{r|}{3.76}      & \textbf{37.45}        & 27.64                 & 34.91                 & 4.12                          \\ \midrule
\multirow{2}{*}{Vicuna-13b-v1.5 \cite{vicuna}}          & Big                   & \textbf{35.82}          & 26.73                      & 37.45                     & \multicolumn{1}{r|}{4.12}      & \textbf{38.00}        & 33.45                 & 28.55                 & 4.67                          \\
                                          & Small                 & \textbf{35.82}          & 30.55                      & 33.64                     & \multicolumn{1}{r|}{2.48}      & \textbf{37.64}        & 34.36                 & 28.00                 & 4.30                          \\
\multirow{2}{*}{Llava-v1.6-Vicuna-13b-hf \cite{llava}} & Big                   & \textbf{47.45}          & 20.18                      & 32.36                     & \multicolumn{1}{r|}{14.12}     & 33.45                 & \textbf{35.27}        & 31.27                 & 1.94                          \\
                                          & Small                 & 32.18                   & 24.91                      & \textbf{42.91}            & \multicolumn{1}{r|}{9.58}      & \textbf{38.91}        & 36.00                 & 25.09                 & 5.58                          \\ \midrule
\multirow{2}{*}{Meta-Llama-3-8B-Instruct \cite{llama3modelcard}} & Big                   & \textbf{36.18}          & 32.91                      & 30.91                     & \multicolumn{1}{r|}{2.85}      & 30.18                 & \textbf{38.18}        & 31.64                 & 4.85                          \\
                                          & Small                 & \textbf{34.18}          & 33.64                      & 32.18                     & \multicolumn{1}{r|}{0.85}      & 30.00                 & \textbf{38.36}        & 31.64                 & 5.03                          \\
\multirow{2}{*}{CogVLM2-llama3-Chat-19B \cite{hong2024cogvlm2}}  & Big                   & \textbf{35.45}          & 30.36                      & 34.18                     & \multicolumn{1}{r|}{2.12}      & \textbf{40.55}        & 33.64                 & 25.82                 & 7.21                          \\
                                          & Small                 & 28.73                   & 32.18                      & \textbf{39.09}            & \multicolumn{1}{r|}{5.76}      & 35.82                 & \textbf{38.91}        & 25.27                 & 5.58                          \\
\multirow{2}{*}{MiniCPM-Llama3-V-2\_5 \cite{yao2024minicpm}}    & Big                   & \textbf{59.45}          & 19.82                      & 20.73                     & \multicolumn{1}{r|}{26.12}     & \textbf{44.55}        & 34.36                 & 21.09                 & 11.21                         \\
                                          & Small                 & 25.09                   & 23.82                      & \textbf{51.09}            & \multicolumn{1}{r|}{17.76}     & \textbf{37.45}        & 28.55                 & 34.00                 & 4.12                          \\ \bottomrule
\end{tabular}}
\caption{Results for Experiment 1 on the unexpanded dataset with semantic size prompt}
\label{tab:exp1_unexpanded_prompt_size}
\end{table}

\end{document}